%% file: iclrV2.0-arxiv.tex
\documentclass{article}
\usepackage{iclr2027_arxiv,times}

\input{math_commands.tex}

\usepackage{hyperref}
\hypersetup{
    hidelinks 
}
\usepackage{url}
\usepackage{graphicx}
\usepackage{caption}
\usepackage{listings}
\usepackage{booktabs}
\usepackage{multirow}
\usepackage{array}
\usepackage{tabularx}
\usepackage{makecell}
\usepackage{subcaption}
\usepackage{xcolor}
\usepackage{pifont}
\usepackage[most]{tcolorbox}
\usepackage{seqsplit}
\usepackage{amsmath}
\usepackage{amssymb}
\usepackage{amsthm}
\usepackage{wrapfig}

\usepackage{booktabs}
\usepackage{longtable}
\usepackage{array}

\definecolor{lightgraybox}{RGB}{245,245,245}
\definecolor{titlebg}{RGB}{228,228,246}
\definecolor{commentgreen}{RGB}{0,140,0}
\definecolor{keywordblue}{RGB}{30,60,200}
\definecolor{stringred}{RGB}{180,40,40}
\definecolor{frameblack}{RGB}{30,30,30}

\definecolor{darkred}{RGB}{170,30,30}
\newcommand{\cmark}{\ding{51}}
\newcommand{\xmark}{\textcolor{darkred}{\ding{55}}}

\lstdefinelanguage{PseudoPython}{
  language=Python,
  morekeywords={while,if,else,for,in,return,True,False,None,and,or,not},
  sensitive=true,
  morecomment=[l]{\#},
  morestring=[b]",
  morestring=[b]'
}

\lstdefinestyle{polycode}{
  language=PseudoPython,
  basicstyle=\ttfamily\footnotesize,
  keywordstyle=\color{keywordblue},
  commentstyle=\color{commentgreen},
  stringstyle=\color{stringred},
  showstringspaces=false,
  breaklines=true,
  breakatwhitespace=false,
  columns=fullflexible,
  keepspaces=true,
  tabsize=2,
  frame=none,
  xleftmargin=0pt,
  xrightmargin=0pt,
  aboveskip=0pt,
  belowskip=0pt
}

\newtcolorbox{polyfigurebox}[1][]{
  enhanced,
  colback=white,
  colframe=frameblack,
  boxrule=0.9pt,
  arc=3pt,
  outer arc=3pt,
  left=8pt,
  right=8pt,
  top=2pt,
  bottom=4pt,
  boxsep=0pt,
  title={\bfseries Demonstrative Code of polyMonitor Pipeline},
  colbacktitle=titlebg,
  coltitle=black,
  fonttitle=\bfseries,
  attach boxed title to top left={xshift=0pt,yshift=0pt},
  boxed title style={
    colframe=frameblack,
    colback=titlebg,
    boxrule=0.9pt,
    arc=0pt,
    left=6pt,
    right=6pt,
    top=3pt,
    bottom=3pt
  },
  #1
}

\usepackage[most]{tcolorbox}
\usepackage{booktabs}
\usepackage{tabularx}
\usepackage{array}
\usepackage{ragged2e}
\usepackage{enumitem}
\usepackage{xcolor}

\definecolor{PromptHeader}{HTML}{B9B9B9}
\definecolor{PromptBorder}{HTML}{C8C8C8}
\definecolor{PromptBody}{HTML}{FCFCFC}

\newcolumntype{Y}{>{\RaggedRight\arraybackslash}X}
\newcolumntype{P}[1]{>{\RaggedRight\arraybackslash}p{#1}}

\newtcolorbox{promptbox}[1]{
    enhanced,
    breakable,
    colback=PromptBody,
    colframe=PromptBorder,
    colbacktitle=PromptHeader,
    coltitle=white,
    fonttitle=\bfseries\large,
    title={#1},
    arc=2mm,
    outer arc=2mm,
    boxrule=0.7pt,
    left=3.5mm,
    right=3.5mm,
    top=2.5mm,
    bottom=2.5mm,
    toptitle=1.6mm,
    bottomtitle=1.6mm,
    before skip=8pt,
    after skip=8pt
}

\usepackage{booktabs}
\usepackage{tabularx}
\usepackage{array}
\usepackage{ragged2e}

\newcommand{\MiniTable}[4]{%
\begin{minipage}[t]{0.318\textwidth}
\vspace{0pt}
\centering
\textbf{#1}\par\vspace{1mm}
{\fontsize{7.2}{8.3}\selectfont
\begin{tabularx}{\linewidth}{
    @{}
    >{\RaggedRight\arraybackslash}p{0.45\linewidth}
    >{\RaggedRight\arraybackslash}X
    @{}
}
\toprule
\textbf{#2} & \textbf{#3} \\
\midrule
#4
\bottomrule
\end{tabularx}}
\end{minipage}%
}

\title{Unlocking the Forecasting Economy: A Suite of Datasets For The Full-Lifecycle of Prediction Markets}

\author{
\textbf{Huaiyu Jia}$^{1,*}$ \quad
\textbf{Luofeng Zhou}$^{2,*}$ \quad
\textbf{Wentao Zhang}$^{3}$ \quad
\textbf{Lin William Cong}$^{3}$ \quad
\textbf{Siguang Li}$^{1}$ \quad
\textbf{Shuo Sun}$^{1}$ \\[3pt]
$^{1}$The Hong Kong University of Science and Technology (Guangzhou) \\
$^{2}$New York University \qquad
$^{3}$Nanyang Technological University \\
{\small $^{*}$Equal contribution}
}

\iclrfinalcopy
\begin{document}

\maketitle

\begin{abstract}
Prediction markets are markets for trading claims on universal future events (e.g., presidential elections). Fueled by a meteoric surge with over \$50 billion trading volume, they have emerged as a promising forecasting mechanism, where their prices provide continuously updated signals of collective beliefs. In decentralized platforms (e.g., Polymarket), the prediction market lifecycle include six stages: market creation, token registration, trading, oracle interaction, dispute, and final settlement. However, comprehensively tracking this complete pipeline remains a major challenge, as the underlying data are severely fragmented across heterogeneous on-chain smart contracts and off-chain sources. To fill this critical gap, we present the first continuously synchronized dataset suite for the full-lifecycle of decentralized prediction markets. To achieve large-scale cross-source integration, incomplete linkage, and continuous synchronization, we build a unified relational data system that integrates three canonical layers: i) market metadata, ii) fill-level trading records, iii) oracle-resolution events, through identifier resolution, on-chain recovery, and incremental updates. The resulting dataset spans from October 2020 to update-to-date and comprise more than 3.29 million market records, over 1.90 billion order execution records, and nearly 21 million oracle events. We describe the data model, collection pipeline, and consistency mechanisms that make the datasets reproducible and extensible. We further demonstrate its utility for multiple communities through NBA outcome calibration for sport traders, CPI expectation reconstruction for economists, and oracle-risk analysis for blockchain researchers. A public website with dataset access, interactive visualizations, and lightweight LLM-assisted exploration tools are publicly available at \textcolor{blue}{\url{https://www.polymonitor.club/}}.
\end{abstract}

\section{Introduction}

Prediction markets trade event-contingent contracts whose prices aggregate collective beliefs about future outcomes \citep{rahman2025sok}. Their market scope spans diverse domains, including elections \citep{cutting2025betting}, sports \citep{reichenbach2025exploring}, macroeconomic releases \citep{diercks2026kalshi} and geopolitics \citep{becker2026political}. Driven by widespread public engagement, prediction market platforms have attracted a rapidly expanding global user base, commenting them as a mainstream crowd-forecasting medium. Specifically, Polymarket serves as the premier operational exemplar in this domain, capturing over 60\% of active trader liquidity and settlement traffic.

\begin{figure}[t]
\centering
\includegraphics[width=0.9\linewidth]{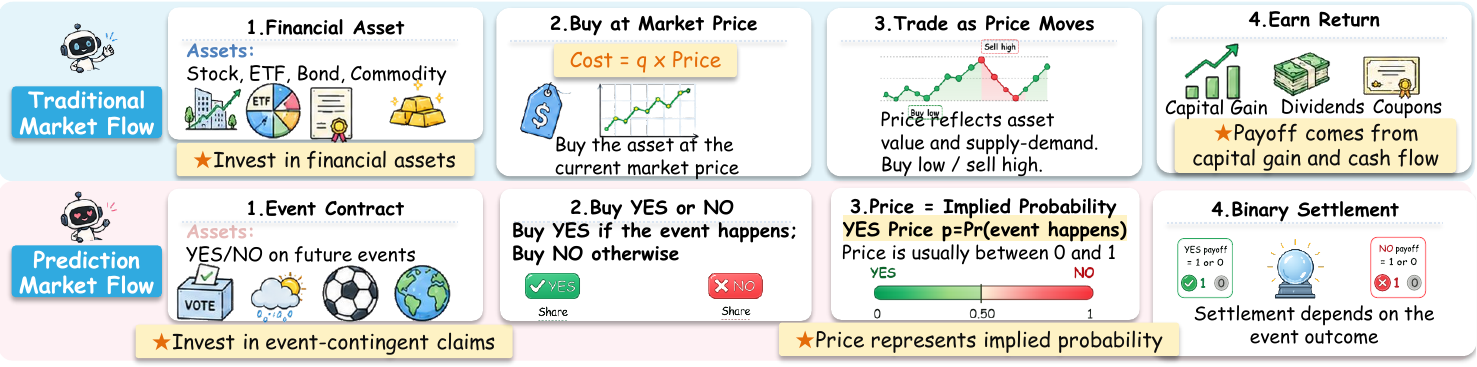}
\caption{Mechanisms comparison between traditional markets and prediction markets}
\label{fig:polymarket_mechanism}
\end{figure}

However, systematic research in this field is fundamentally constrained by the lack of high-quality unified datasets despite the great potential. Existing work either focuses on narrow settings (e.g., arbitrage \citep{saguillo2025unravelling} and price discovery \citep{ng2026price}) or isolated phenomena (e.g., macro markets \citep{diercks2026kalshi} and US president election. As illustrated in Figure~\ref{fig:polymarket_mechanism}, traditional markets trade financial assets where price fluctuation reflect investment return. In contrast, prediction markets trade event-contingent YES/NO claims where prices lies in $[0,1]$ and settle to binary payoffs after the event outcome is determined. Due to the complexity of the mechanism, it is non-trivial to build comprehensive datasets with several major challenges. First, heterogeneous data sources make the representation of one market different across off-chain APIs, exchange contracts, and oracle contracts, requiring robust cross-source entity resolution. Second, transaction logs and oracle events do not directly expose all market-level semantics so that missing links need to be recovered auxiliary mappings and on-chain registration records. Third, it is neccessary for high-quality datasets in this domain to support real-time synchronization with a live ecosystem.

In this work, we present the first suite of datasets to systematically capture the full-lifecycle of decentralized prediction markets at scale from market creation, on-chain trading to oracle resolution and settlement. The pipeline supports both historical back-
filling and continuous updates, and integrates three linked layers: market metadata defining each
contract, OrderFilled records capturing fill-level executions and price evolution, and oracle events documenting requests, proposals, disputes, and final settlements. We develop a unified relational design with canonical market, trade, and oracle relations, together with bridge tables, cache layers, and synchronization metadata that make ingestion resumable, duplicate-safe, and extensible. Our dataset spans October 2020 to September 2026 and contains more than 3.29 million market records, over 1.90 billion trade fill records, and nearly 20 million oracle events. Beyond dataset construction, we conduct ecosystem-scale descriptive analyses to characterize market coverage, topic composition, platform growth, trading activity, wallet participation, and
oracle-resolution dynamics. We further demonstrate the utility of the dataset through representative use cases for different communities: NBA probability calibration and market analysis for traders, CPI expectation reconstruction for economists and resolution-aware supervised learning from partial trading histories for the machine learning community. Finally, our public website provides dataset access, interactive visualizations, and lightweight LLM-assisted tools for exploring markets and querying lifecycle data.

\section{Data Collection}
\subsection{Challenges and our Approaches}
Constructing a full-lifecycle dataset for decentralized prediction markets requires more than collecting records from individual sources. The central difficulty is to recover a temporally coherent market process from data that differ in semantics, identifiers, scale, and arrival time. We therefore formulate the construction problem in terms of five design challenges:

\begin{itemize}
\item \textbf{C1:} How should the lifecycle boundary of a prediction market be defined?
\item \textbf{C2:} How can all relevant market, trading, oracle, and settlement records be discovered?
\item \textbf{C3:} How can heterogeneous records be linked when they expose different identifiers?
\item \textbf{C4:} How should large lifecycle records be stored and queried efficiently?
\item \textbf{C5:} How can the dataset remain continuously updated without sacrificing reproducibility?
\end{itemize}

\begin{figure}
    \centering
    \includegraphics[width=0.99\linewidth]{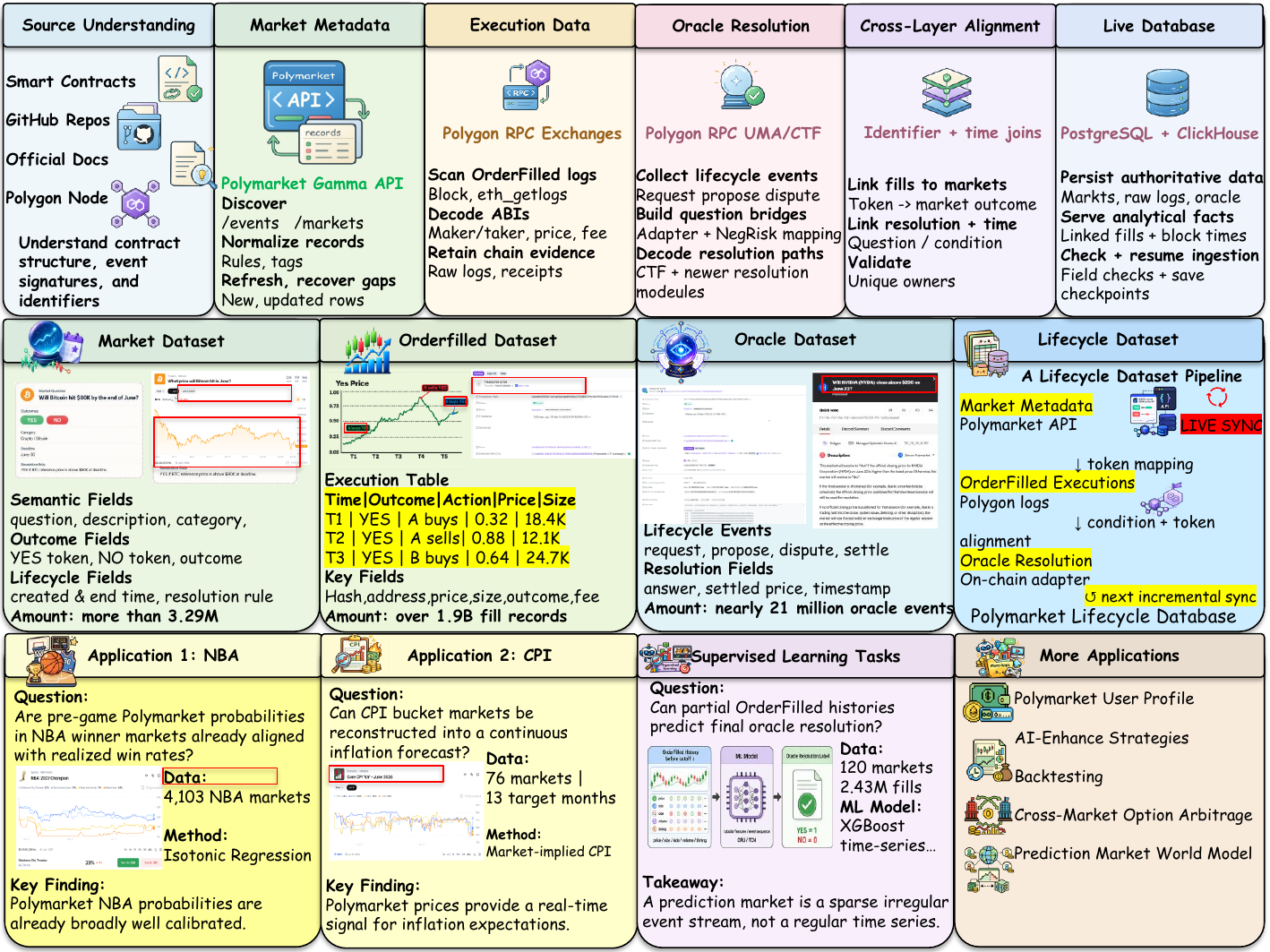}
    \caption{Overview of the Polymarket lifecycle dataset pipeline and representative applications.}
    \label{fig:dataset_overview}
\end{figure}

Figure~\ref{fig:dataset_overview} summarizes the online construction pipeline. Since no single source provides a complete market history, we organize dataset construction around five questions concerning lifecycle definition, data discovery, cross-source alignment, storage, and continuous synchronization.

\paragraph{A1: }For market $m$, let $\mathcal{M}_m$ denote its metadata, $\mathcal{F}_m$ its OrderFilled executions, and $\mathcal{O}_m$ its oracle events. We define the lifecycle at snapshot time $T$ as

\[
\mathcal{L}_m(T)=
\left(
\mathcal{M}_m,\,
\{f\in\mathcal{F}_m:\tau_m^{s}\leq t_f\leq\tau_m^{e}(T)\},\,
\{o\in\mathcal{O}_m:\tau_m^{s}\leq t_o\leq\tau_m^{e}(T)\}
\right),
\]

\[
\tau_m^{s}=\min\left(t_m^{\mathrm{created}},t_m^{\mathrm{registered}},
\min_{f\in\mathcal{F}_m}t_f\right),
\qquad
\tau_m^{e}(T)=
\begin{cases}
t_m^{\mathrm{settle}}, & \text{if settled},\\
T, & \text{otherwise}.
\end{cases}
\]

This definition includes delayed metadata and right-censored live markets. A lifecycle becomes complete when the market definition, execution history, and final settlement are all observed.

\paragraph{A2:}
Constructing a full-lifecycle Polymarket dataset is challenging because no single source exposes the complete market process: relevant records are fragmented across platform APIs \citep{polymarketGammaAPI2026}, smart-contract code \citep{polymarketContracts2026}, public repositories \citep{polymarketGitHub2026}, Polygon logs \citep{polygonRPC2026}, and UMA oracle contracts \citep{uma_protocol}, each with its own schema, identifier space, and update cadence . We therefore first map Polymarket's observable data surface by jointly analyzing contract implementations, official documentation, API responses, on-chain transactions, and oracle architectures. This mapping identifies the deployed contracts, relevant event families, and cross-system identifier mappings required to link market definitions, trading activity, and final outcomes.

Market metadata---questions, deadlines, categories, resolution rules, condition identifiers, and outcome tokens---are retrieved from the Gamma and CLOB interfaces \citep{polymarketTradingAPI2026}, with markets missing or delayed in these APIs recovered from on-chain token-registration records. Trading data are collected independently by scanning Polygon logs across all supported Polymarket exchange deployments and fill-event formats, spanning both legacy and current market designs; this execution-first strategy prevents trades from being omitted when metadata are incomplete or arrive late. Oracle histories are reconstructed from UMA contracts and Polymarket resolution adapters  \citep{polymarketUmaCtfAdapter2026} by recovering the complete resolution sequence, comprising market initialization, outcome requests, proposals, disputes, and final settlement. Together, these three components constitute our lifecycle dataset.

\paragraph{A3:}Each contract is assigned a canonical market identifier while preserving its original Gamma market identifier, condition identifier, question identifier, and outcome-token identifiers. Execution records are linked through the traded outcome token:

\[
\text{traded token}
\rightarrow
\text{outcome-token mapping}
\rightarrow
\text{canonical market}.
\]

This mapping connects each execution to its corresponding contract and determines whether the traded token represents the YES or NO outcome.

Oracle records are matched using shared question and condition identifiers extracted from UMA and adapter data, with normalized titles and resolution deadlines used only when direct identifier links are unavailable. This procedure groups market metadata, executions, and oracle outcomes under the same canonical market identifier.

\paragraph{A4:}
We use PostgreSQL and ClickHouse because the three data layers have different storage and query requirements. PostgreSQL stores market metadata and oracle records, which are relatively compact, relational, and frequently enriched through identifier mappings and lifecycle updates. ClickHouse stores the much larger OrderFilled execution stream, whose append-oriented structure benefits from columnar compression, fast block-range scans, deterministic event ordering, and efficient market-, transaction-, and address-level queries. This separation allows PostgreSQL to manage relational lifecycle state while ClickHouse supports large-scale execution analysis.

\paragraph{A5:}
The dataset is maintained through a checkpoint-based synchronization pipeline that continuously collects new market metadata, Polygon executions, and oracle-resolution events. Each synchronization round revisits a small overlap window to recover delayed responses, temporary RPC failures, and short chain reorganizations, while idempotent insertion prevents duplicate records. Incomplete ranges are retained for later repair rather than silently discarded. We also maintain a scheduled Kaggle update script that periodically exports refreshed dataset snapshots and accompanying coverage metadata, enabling users to access continuously updated data.

\subsection{Dataset Composition}

Our dataset organizes Polymarket into three linked lifecycle layers: market definition, trading, and resolution. At the fixed cutoff of September 1, 2026, 00:00 UTC (last included Polygon block: 93,010,514), the snapshot contains \textbf{3,290,907} consolidated market and protocol-condition identities, \textbf{1,901,194,769} deduplicated OrderFilled events, and \textbf{21,028,317} oracle-, adapter-, and protocol-side lifecycle events. Across the catalog, {2,441,701 identities (74.20\%)} have observed OrderFilled activity, {3,077,506 (93.52\%)} have verified final-resolution evidence, and {2,402,991 (73.02\%)} contain both, yielding more than {2.4 million linked trading-to-outcome lifecycles}. The remaining coverage differences primarily arise from heterogeneous market lifecycles and source availability: some markets predate reliably recoverable trading records or have been removed from current platform interfaces, some had not reached final settlement by the snapshot cutoff and therefore lack oracle outcomes, and a small subset lacks both trading and resolution evidence because the corresponding official metadata are incomplete or ambiguous. More details in the Appendix \ref{data_statistics}

\textbf{Market dataset.}
The current snapshot contains \textbf{3,290,907} consolidated market and protocol-condition identities, providing broad coverage across the heterogeneous market structures and protocol generations that have appeared throughout Polymarket's history. This scale makes the market layer useful not only as a contract catalog, but also as a foundation for cross-market comparison, event-level grouping, semantic analysis, and downstream applications.

\textbf{OrderFilled dataset.}
The execution layer contains \textbf{1,901,194,769} deduplicated on-chain OrderFilled logs with transaction hash, block number, timestamp, participant addresses, token, side, price, size, and fee. Of these, \textbf{1,901,194,698} are linked to identities in the market catalog, leaving only 71 records associated with 21 genuine combinatorial conditions not yet represented in the catalog snapshot. In total, OrderFilled activity covers {2,441,701 catalog identities (74.20\%)}, supporting analysis at market, token, block, transaction, and address granularity. 

\textbf{Oracle dataset.}
The resolution layer contains \textbf{21,028,317} lifecycle events spanning multiple generations of Polymarket's oracle and settlement infrastructure. Major components include {3,338,899} CTF condition-preparation events, {3,025,824} CTF final-result events, {2,340,129} UMA requests, {2,123,464} proposals, {4,334} disputes, {2,122,993} UMA settlements, {2,333,185} adapter initializations, and {2,110,388} {QuestionResolved} events, together with newer protocol-module events.

\subsection{demonstrative example : The 2024 U.S. Presidential Election}

\begin{figure}[t]
    \centering
    \includegraphics[width=\linewidth]{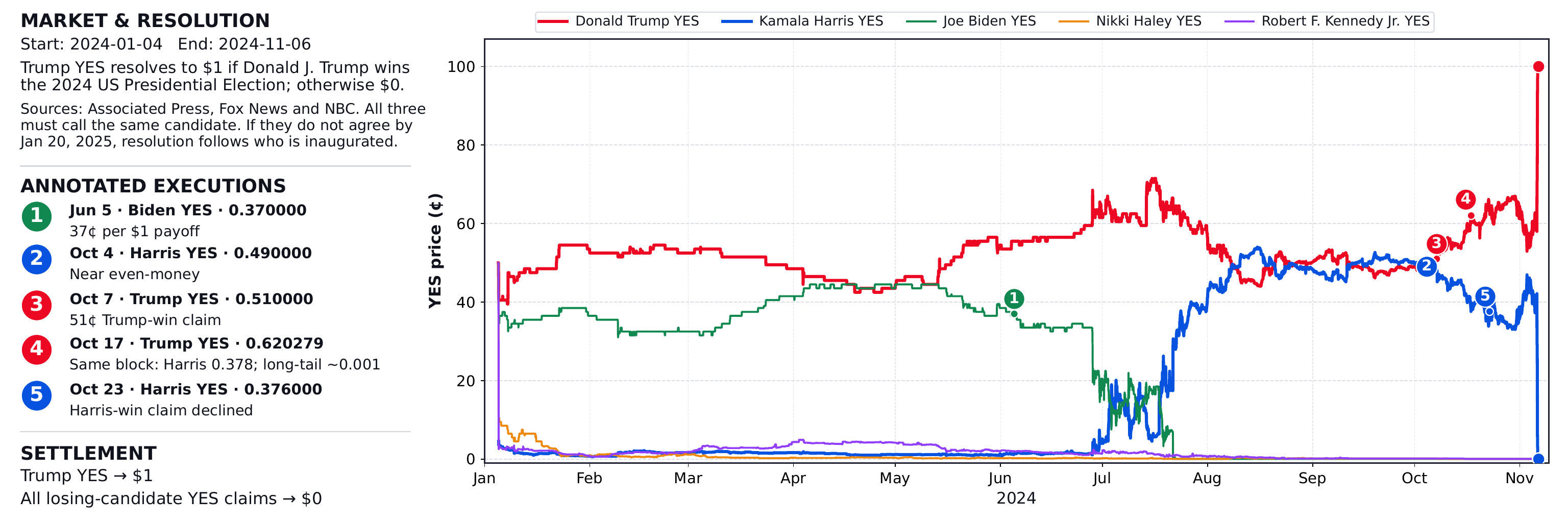}
    \caption{
    Full-lifecycle reconstruction of the 2024 U.S. Presidential Election market family. 
    Candidate-level YES-price trajectories are reconstructed from OrderFilled executions and linked to their final oracle settlements; the winning contract settles at one, while losing settle at zero.
    }
    \label{fig:lifecycle}
\end{figure}




Figure~\ref{fig:lifecycle} traces how the three layers connect for the 2024 U.S. Presidential Election market family. Polymarket listed the election not as a single multinomial contract but as a set of linked candidate-level binary markets, with separate YES/NO contracts for Donald Trump, Kamala Harris, Joe Biden, Nikki Haley, and Robert F. Kennedy Jr.; the market layer preserves their shared election-level context together with the identifiers, outcome tokens, trading windows, and resolution rules of each candidate contract. The OrderFilled layer then supplies an execution-based YES-price trajectory per contract, derived solely from fills: at an observed snapshot on October 17, 2024, the latest YES prices stood at 0.620279 for Trump, 0.378000 for Harris, and roughly 0.001000 for the long-tail candidates, and over the full lifecycle the trajectories track the Biden contract's decline, Harris's subsequent rise, and the late Trump--Harris divergence. After the election result became known, the oracle layer closed the lifecycle, settling the Trump contract at 1.0 and the Biden, Harris, Haley, and Kennedy contracts at 0.0. Joining these settlements to the market definitions and execution trajectories completes a per-contract record from creation and price discovery to realized outcome.

\section{downsteam applications}

The preceding analyses characterize the structure, dynamics, and resolution process of our Polymarket lifecycle dataset. To demonstrate its value, we further evaluate four representative use cases:

Q1: For \textbf{traders} specializing in NBA prediction markets, our dataset supports the reconstruction of market-implied probabilities and trading-signal evaluation.

Q2: For \textbf{macroeconomists and policy researchers}, our dataset enables systematic evaluation of whether prediction-market prices anticipate CPI releases and provide predictive information beyond conventional economic indicators.

Q3: For \textbf{machine-learning researchers}, partial {OrderFilled} histories aligned with oracle outcomes provide a large-scale supervised benchmark for early resolution prediction and temporal modeling.

Q4: For \textbf{multi-agent system developers and prediction-market platform builders}, our lifecycle dataset can ground LLM-based multi-agent systems that jointly perform market discovery, execution analysis, risk monitoring, and resolution tracking with auditable evidence.

All experiments are conducted using our released datasets. Together, these use cases demonstrate that the lifecycle data support both domain-specific decision making and general predictive modeling, rather than serving only as a descriptive record of market activity.

\subsection{Market-Aware Trading with SportsNBA}

\begin{wrapfigure}{r}{0.45\textwidth}
    \centering
    \includegraphics[width=\linewidth]{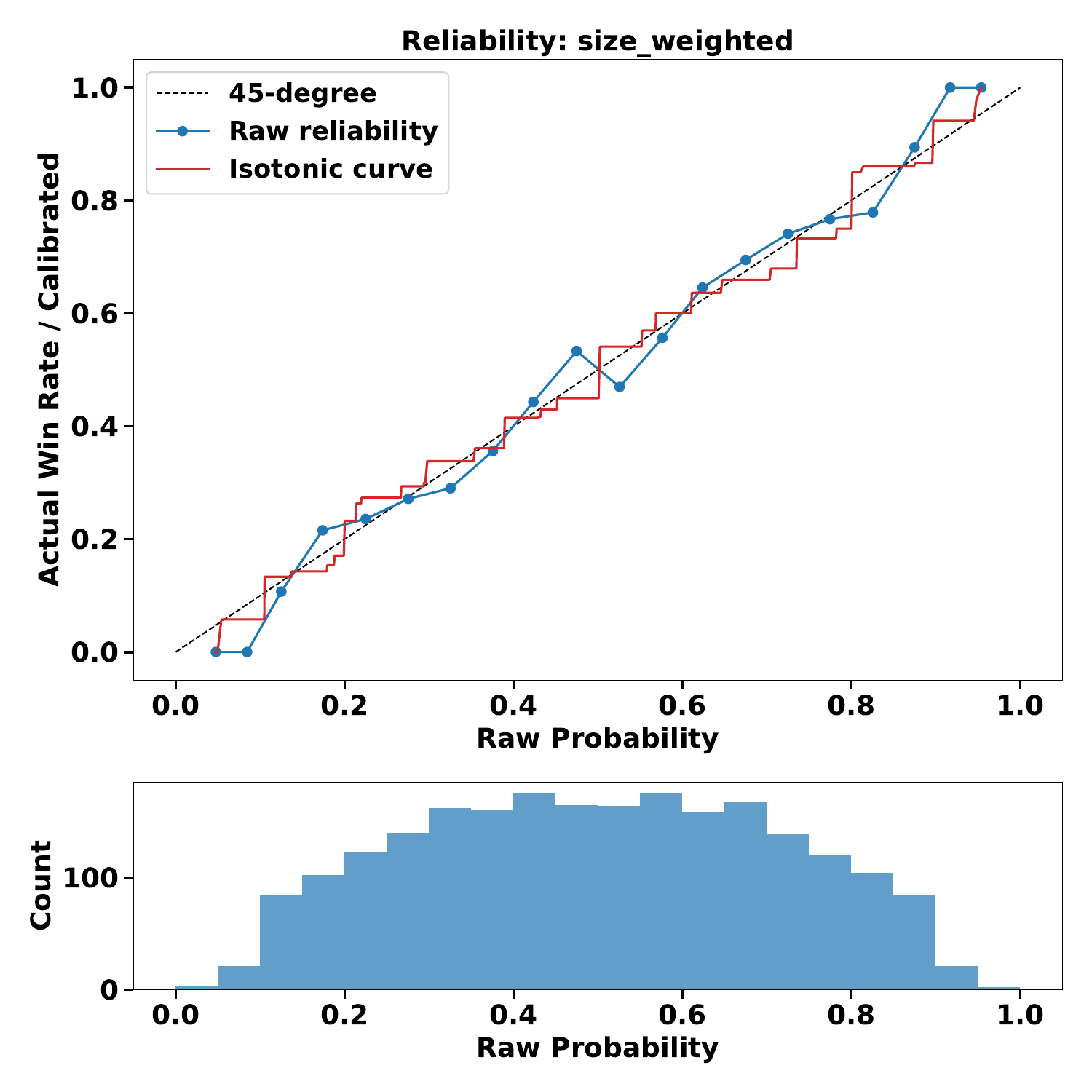}
    \caption{
Calibration of trade-size-weighted pre-game probabilities in NBA winner markets. Raw market probabilities closely track realized win frequencies, while isotonic calibration provides little improvement.
}
    \label{fig:sizeWeighted}
\end{wrapfigure}

A trader holding a view on an NBA game must first know whether the market's own pre-game price is already an accurate probability: if it is, only a genuine edge over that price can justify a trade, and any backtest must measure deviations from it rather than from a stale or leaky estimate. We therefore use the SportsNBA subset to test whether Polymarket executions support leakage-free, point-in-time fair-value estimation. We retain single-game binary winner contracts and join market metadata, pre-game OrderFilled executions, and oracle outcomes; after removing markets whose game-time alignment is unreliable, the experiment contains 3,869 markets. For each team, the pre-game market probability is the trade-size-weighted average of valid execution prices observed before a chosen cutoff:
\[
p_{\mathrm{pre}}
=
\frac{\sum_{i:t_i<t_{\mathrm{start}}} s_i p_i}
{\sum_{i:t_i<t_{\mathrm{start}}} s_i},
\]
where \(p_i\) and \(s_i\) denote the price and size of execution \(i\) and \(t_{\mathrm{start}}\) is the tip-off time. Reconstructing this quantity separately at 24, 6, and 1 hours before tip-off, using only executions observed before each cutoff, yields a point-in-time price series that never incorporates later information.

We train on games before the 2026 season and reserve the 2026 season for testing, yielding 1,627 training and 1,136 test markets. Against a one-dimensional isotonic calibration model \citep{barlow1972isotonic}, the raw market probability is already well calibrated: the reliability curve in Figure~\ref{fig:sizeWeighted} stays close to the diagonal, and isotonic recalibration brings little improvement. On the test set the raw probabilities achieve a Brier score of 0.2034, a LogLoss of 0.5918, an ECE of 0.0275, and an MCE of 0.0845. The point-in-time reconstruction further shows that prices sharpen as tip-off approaches: on the 2026 test season the Brier score falls from 0.2047 at T-24h to 0.1999 at T-1h while AUROC rises from 0.7410 to 0.7525. Pre-game prices thus behave as an efficient, progressively informative consensus rather than a biased quote.

Finally, we ask whether executable signals exist on top of this consensus. At T-6h we train a residual model over recent price moves, volume, order flow, liquidity, and wallet activity that learns only an adjustment to the current market probability; models are trained on early-2025 markets, selected on late-2025 markets, and frozen before 2026. The residual model does not improve probability accuracy, and under execution-aware scoring---subsequent observed fills, a 60-second delay, limited participation, and a one-cent-per-share cost---the validation-period ROI is positive but its 95\% confidence interval includes zero, while the frozen strategy loses money in 2026. We therefore find no persistent alpha in the tested features. The dataset's value here is instead infrastructural: leakage-free fair-value estimates, cost-aware execution evaluation, and out-of-sample tests that separate robust signals from overfitting, supporting signal development, no-trade decisions, and risk control even when a strategy fails to beat the market.

\begin{figure*}[t]
    \centering
    \captionsetup{font=small}

    \begin{subfigure}[t]{0.32\textwidth}
        \centering
        \includegraphics[width=\linewidth,trim=5 5 5 5,clip]{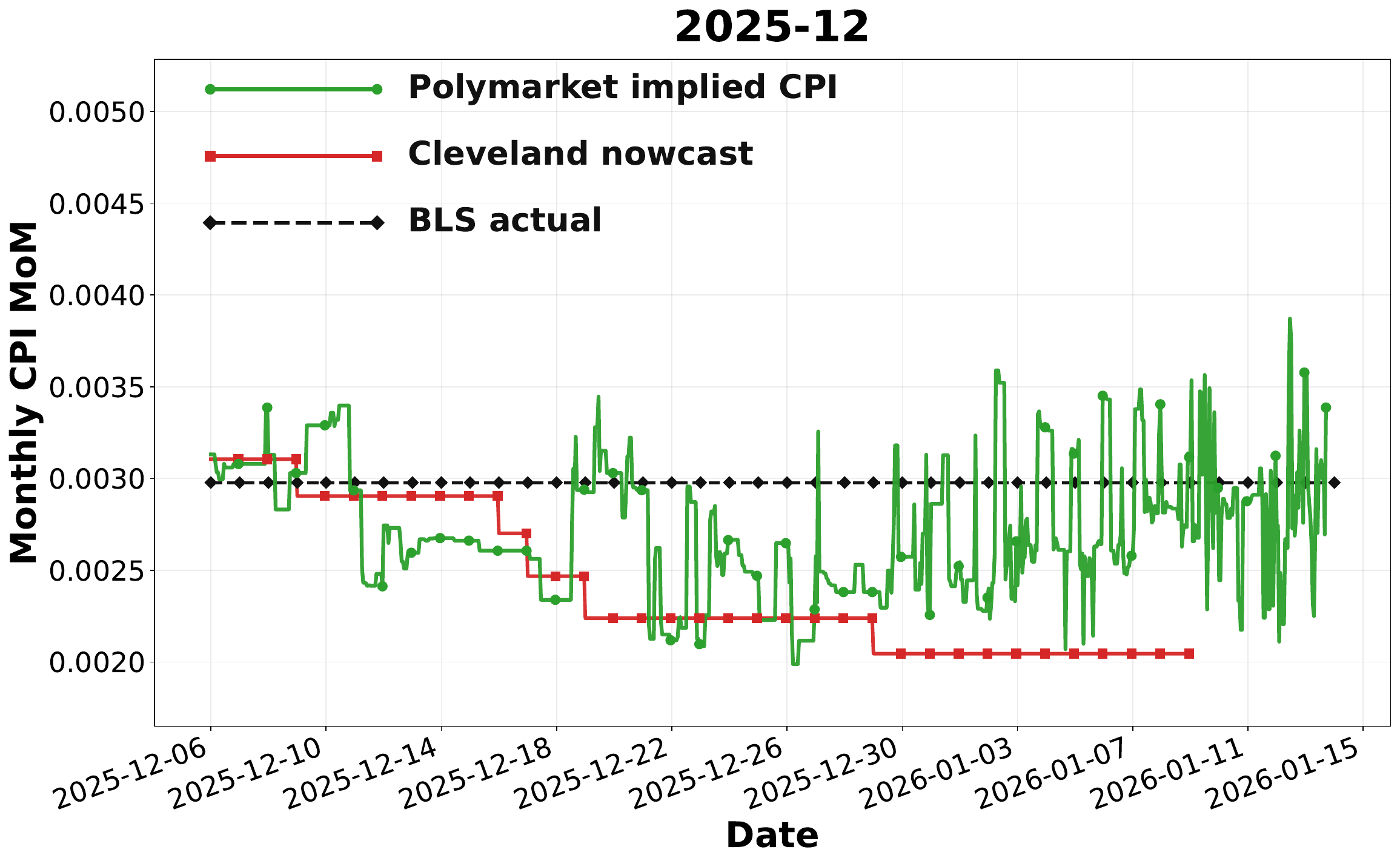}
        \caption{Target month 2025-12.}
        \label{fig:cpi_size_weighted_2025_12}
    \end{subfigure}
    \hfill
    \begin{subfigure}[t]{0.32\textwidth}
        \centering
        \includegraphics[width=\linewidth,trim=5 5 5 5,clip]{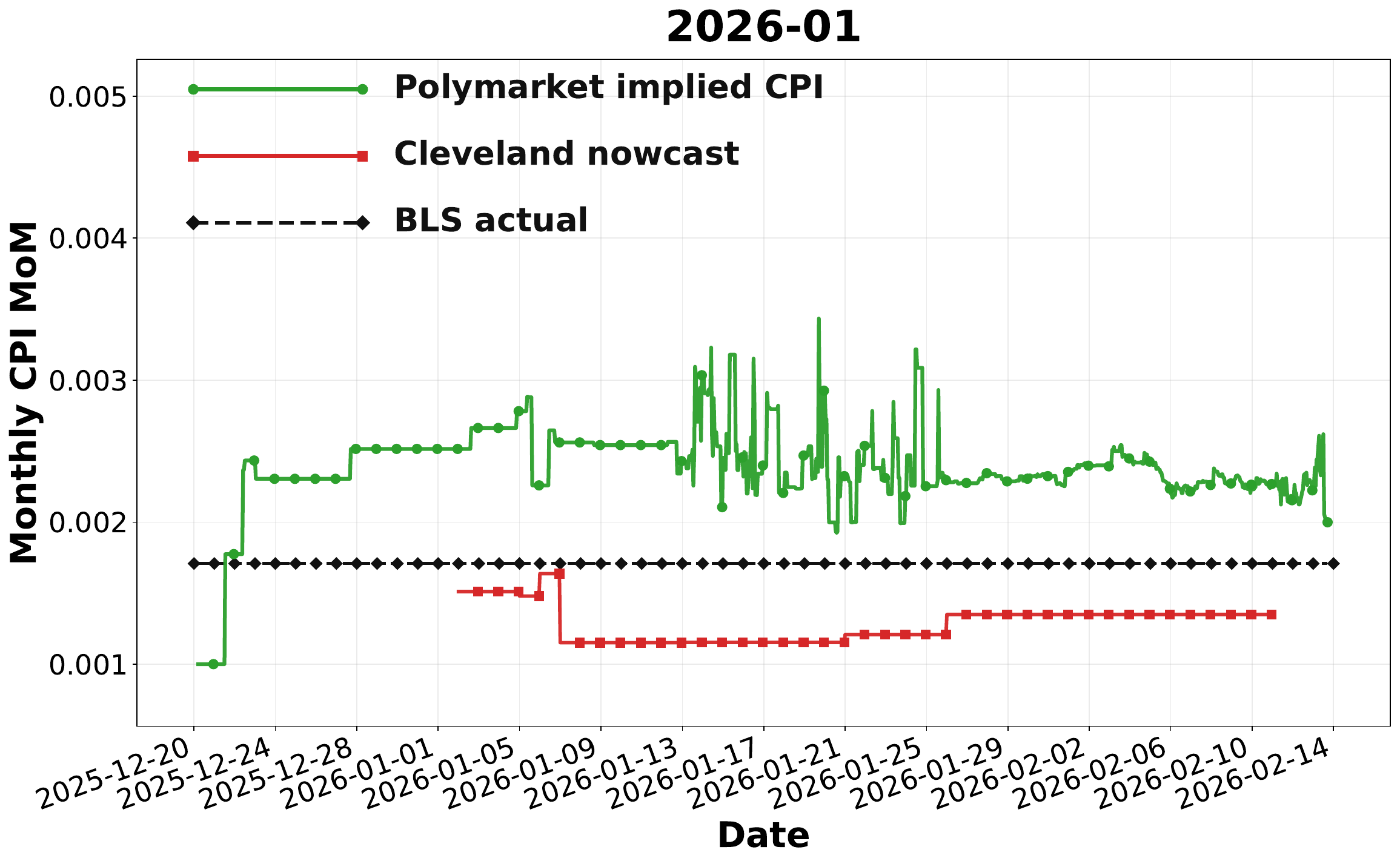}
        \caption{Target month 2026-01.}
        \label{fig:cpi_size_weighted_2026_01}
    \end{subfigure}
    \hfill
    \begin{subfigure}[t]{0.32\textwidth}
        \centering
        \includegraphics[width=\linewidth,trim=5 5 5 5,clip]{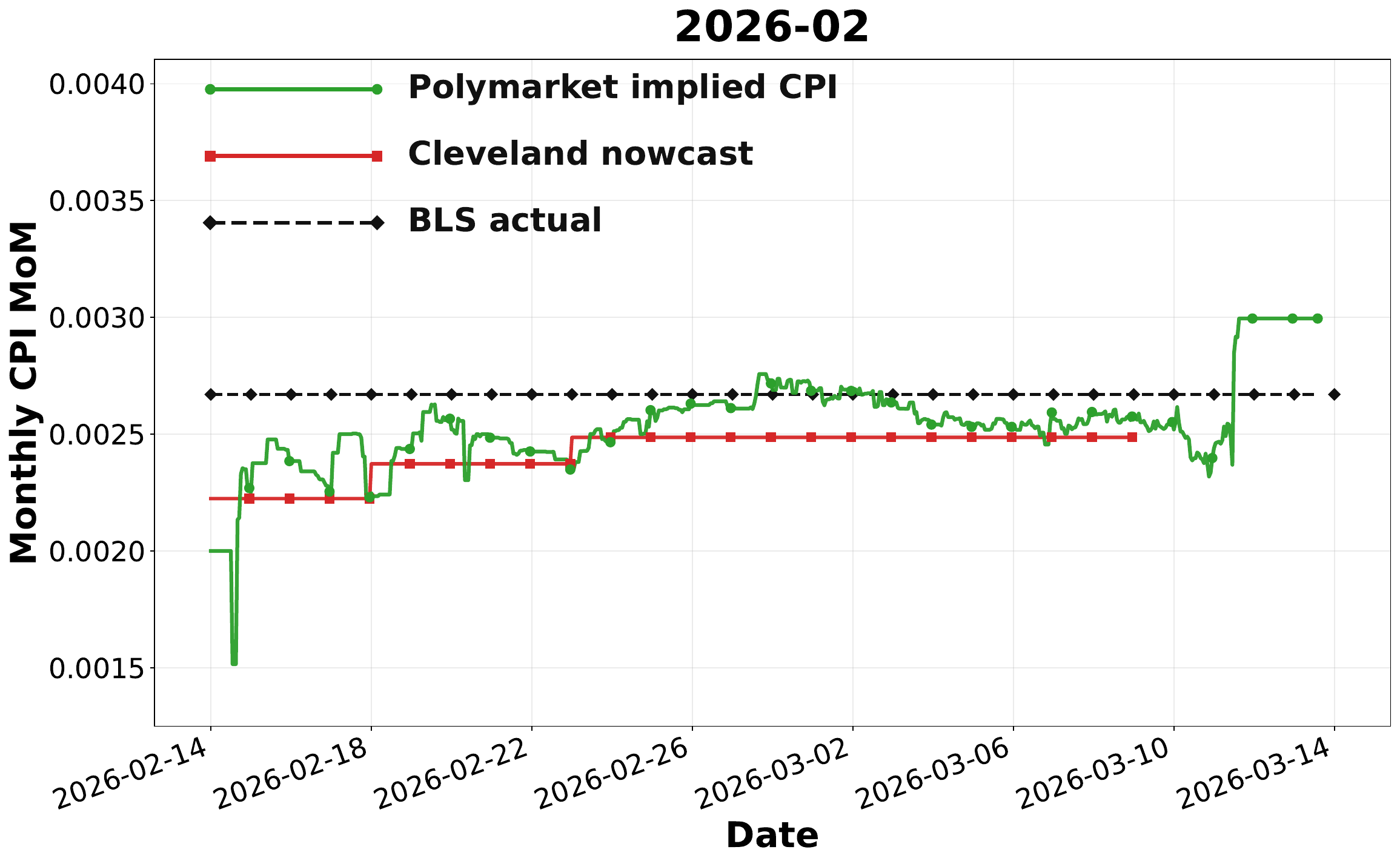}
        \caption{Target month 2026-02.}
        \label{fig:cpi_size_weighted_2026_02}
    \end{subfigure}

    \caption{Polymarket-implied CPI versus Cleveland Fed nowcast and realized BLS CPI.}
    \label{fig:cpi_size_weighted_three_months}
\end{figure*}

\subsection{Market-Implied CPI Expectations}

The Consumer Price Index (CPI) is among the most closely watched macroeconomic indicators, yet on Polymarket it is not traded as a point forecast: each target month lists mutually exclusive bucket markets---e.g., monthly inflation of ``0.1\%,'' ``0.2\%,'' or ``0.4\% or more''---whose prices encode categorical event probabilities rather than a numerical forecast. Comparing market beliefs with macroeconomic statistics therefore first requires reconstructing a continuous implied CPI from these fragmented prices, which we then contrast with the Cleveland Fed nowcast \citep{clevelandfed_inflation_nowcast} and the final BLS CPI \citep{bls_cpi_release_schedule}.

The reconstruction proceeds in three steps. We first unify event semantics: for a trade on a given bucket, a YES execution is taken at its transaction price as the probability of that bucket, while a NO execution is mapped to the same bucket through \(1-\text{price}\). We then summarize beliefs at each timestamp \(t\) with a rolling 24-hour window, computing for each bucket token the trade-value-weighted probability
\[
\hat{p}_{i,t}=\frac{\sum_{k \in \mathcal{T}_{i,t}} v_k p_k}{\sum_{k \in \mathcal{T}_{i,t}} v_k},
\]
where \(\mathcal{T}_{i,t}\) is the set of trades on token \(i\) inside the window, \(p_k\) is the unified event probability of trade \(k\), and \(v_k\) is its trade value; tokens with \(\hat{p}_{i,t} \leq 0.10\) are dropped so that inactive or near-zero buckets do not distort the fit.

Finally, we treat the cross-bucket probability masses at time \(t\) as observations of a unimodal Gaussian over monthly CPI and fit its mean \(\mu_t\) and standard deviation \(\sigma_t\) by matching bucket interval masses,
\[
P_i(\mu_t,\sigma_t)=
\Phi\!\left(\frac{upper_i-\mu_t}{\sigma_t}\right)
-
\Phi\!\left(\frac{lower_i-\mu_t}{\sigma_t}\right),
\]
where \(\Phi(\cdot)\) is the standard normal CDF and \([lower_i, upper_i]\) is the interval of bucket \(i\). Fitting interval masses rather than raw token prices respects the categorical structure of the contracts, and the fitted mean \(\mu_t\) serves as the market-implied monthly CPI at time \(t\). After cleaning and alignment, the experiment covers 76 CPI markets, 150,981 executions, 65,496 panel rows, and 13 target months with released and comparable BLS outcomes.

Figure~\ref{fig:cpi_size_weighted_three_months} compares the reconstructed series with both benchmarks over three representative months. The market-implied path tracks the realized BLS value more closely than the Cleveland nowcast in 2025-12 and 2026-02, whereas in 2026-01 both series miss the final print in opposite directions. The market-based estimate is therefore not uniformly superior; its value lies in continuous updating as trades arrive, offering a high-frequency complement to smoother expectations and exposing where tradable inflation beliefs diverge from the professional consensus.

\subsection{Supervised Learning Task}
\label{sec:supervised-learning}

\textbf{Resolution prediction from partial market histories.}
We formulate a supervised benchmark that predicts the final resolution of a binary prediction market from information observed before settlement. For market \(m\) at lifecycle cutoff \(\tau\), the model receives its partial execution history and market metadata and outputs the probability of a YES resolution:
\[
\widehat{p}_{m,\tau}
=
f_{\theta}\!\left(
\Phi(\mathcal{H}_{m}^{\leq \tau},z_m)
\right),
\qquad
y_m \in \{0,1\}.
\]
Here, \(\mathcal{H}_{m}^{\leq \tau}\) contains only executions available before the cutoff, \(z_m\) denotes market metadata, and \(y_m\) is obtained from the final oracle settlement. This task exposes a distinctive machine-learning setting: labels are supplied naturally by oracle outcomes, while inputs are sparse, irregular, variable-length event sequences that describe how collective beliefs evolve over time.

\textbf{Data processing and representations.}
The benchmark contains 120 resolved binary markets observed at three lifecycle stages: the first \(20\%\) of trading, the first \(50\%\), and 24 hours before settlement. This produces 360 examples. Markets are split chronologically into 84 training, 18 validation, and 18 test markets, corresponding to 252, 54, and 54 examples, with no market shared across splits.
We evaluate complementary representations of the same lifecycle data. Tabular inputs summarize the latest price, VWAP, volume, volatility, trade count, and remaining time. Regularized sequence inputs aggregate executions into lifecycle-normalized intervals, while event-level inputs preserve the most recent fills together with price, size, side, time gaps, and missingness masks. This design allows researchers to compare feature aggregation, regular time-series modeling, and direct learning from irregular transaction streams under the same labels and splits.

\textbf{Models and experimental setup.}
The benchmark covers classical and recent model families: Random Forest and MLP for tabular features; the open TabICLv2 table foundation model; an event-level GRU; PatchTST and TimesNet from the THUML Time-Series-Library; MiniROCKET; a frozen MantisV2 encoder with a validation-selected linear probe; and Neural CDE for continuous-time irregular sequences.  All experiments were executed in the same conda environment with CUDA 12.8 on a Linux workstation equipped with two AMD EPYC 9554 processors, 1.1 TiB of memory, and an NVIDIA GeForce RTX 5090 with 32 GB of GPU memory. Hyperparameters and model selection use only the training and validation markets; the test split is evaluated once. Performance is measured using Accuracy, AUROC, Log Loss, and Brier score.

\begin{table*}[t]
\centering

\caption{
Resolution prediction on held-out test markets, pooled across three lifecycle
cutoffs. Brackets report 95\% market-level cluster bootstrap confidence
intervals based on 10,000 resamples. AUPRC treats YES as the positive class.
}
\label{tab:supervised-outcome-prediction}

\vspace{2pt}

\resizebox{\textwidth}{!}{%
\begin{tabular}{@{}lccccc@{}}
\toprule
Model
& AUROC $\uparrow$ [95\% CI]
& AUPRC $\uparrow$
& Log Loss $\downarrow$ [95\% CI]
& Brier $\downarrow$ [95\% CI]
& ECE $\downarrow$ \\
\midrule

Last Price
& 0.804 [0.615, 0.942]
& 0.681
& \textbf{0.474} [0.308, 0.655]
& \textbf{0.162} [0.099, 0.231]
& \textbf{0.078} \\

Random Forest \citep{breiman2001random}
& 0.764 [0.582, 0.915]
& 0.642
& 0.522 [0.398, 0.653]
& 0.176 [0.122, 0.233]
& 0.097 \\

MLP \citep{rumelhart1986learning}
& \textbf{0.827} [0.660, 0.956]
& 0.738
& 0.523 [0.434, 0.609]
& 0.173 [0.134, 0.213]
& 0.142 \\

TabICLv2 \citep{qu2026tabiclv2}
& 0.620 [0.346, 0.861]
& 0.493
& 0.779 [0.444, 1.178]
& 0.261 [0.149, 0.391]
& 0.251 \\

Event-level GRU \citep{cho2014learning}
& 0.814 [0.624, 0.969]
& \textbf{0.758}
& 0.588 [0.533, 0.643]
& 0.199 [0.172, 0.226]
& 0.218 \\

PatchTST \citep{nie2022time}
& 0.752 [0.573, 0.900]
& 0.562
& 0.567 [0.440, 0.703]
& 0.193 [0.135, 0.256]
& 0.091 \\

MantisV2 \citep{feofanov2026mantisv2}
& 0.711 [0.451, 0.918]
& 0.511
& 0.651 [0.602, 0.706]
& 0.229 [0.205, 0.256]
& 0.146 \\

MiniROCKET \citep{dempster2021minirocket}
& 0.610 [0.383, 0.802]
& 0.438
& 0.640 [0.543, 0.739]
& 0.225 [0.180, 0.271]
& 0.112 \\

TimesNet \citep{wu2022timesnet}
& 0.576 [0.284, 0.843]
& 0.457
& 0.657 [0.422, 0.947]
& 0.222 [0.130, 0.330]
& 0.105 \\

Neural CDE \citep{kidger2020neural}
& 0.453 [0.296, 0.619]
& 0.311
& 0.844 [0.717, 0.965]
& 0.320 [0.260, 0.377]
& 0.307 \\

\bottomrule
\end{tabular}%
}

\end{table*}

\textbf{Findings.}
The event-level GRU attains the highest Accuracy and the MLP the highest AUROC, evidence that raw execution sequences and aggregated trading features carry complementary signal for outcome classification and ranking. Yet the latest market price stays clearly ahead in Log Loss and Brier score: the market's own quote remains a probabilistic baseline that none of these models surpass. The benchmark thus separates two objectives often conflated---picking the most likely outcome versus producing a calibrated resolution probability---and provides a reproducible testbed for tabular learning, irregular event-sequence modeling, and probability calibration on naturally aligned market histories and oracle labels.

\begin{figure*}[t]
    \centering
    \includegraphics[width=\textwidth]{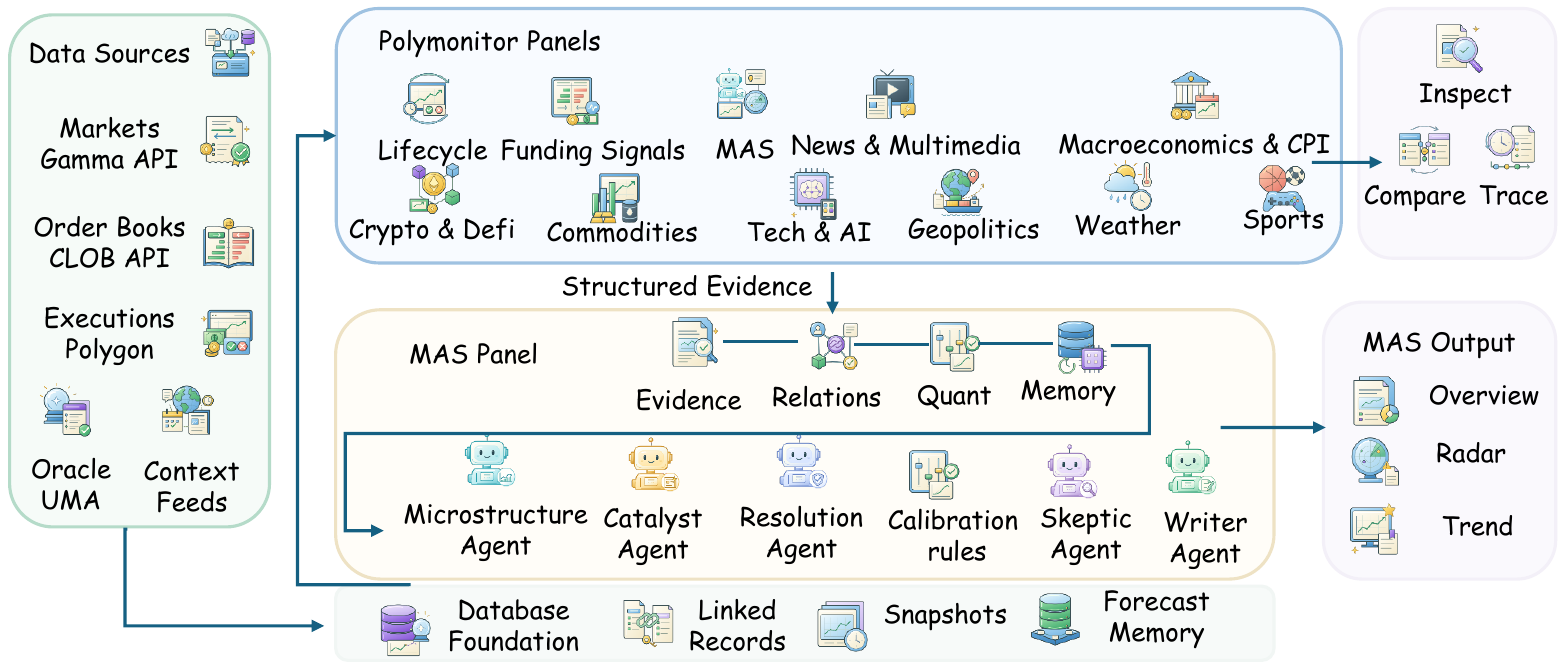}
    \caption{
    Architecture of the Polymonitor and multi-agent lifecycle system.
    Heterogeneous market, execution, oracle, and external-context data
    are processed by specialized agents and unified through shared
    lifecycle state, memory, and analytical tools. The resulting outputs
    support market monitoring, execution analysis, risk assessment,
    resolution tracking, and archival research.
    }
    \label{fig:polymonitor_mas}
\end{figure*}

\subsection{Polymonitor: A Lifecycle-Oriented Research Platform}
\label{app:polymonitor}

Polymonitor turns the dataset into an interactive research environment organized around each contract's lifecycle. It joins market and event metadata (Gamma API), order-book states (CLOB API), Polygon executions, UMA oracle events, and external context feeds on persistent market, token, condition, and question identifiers (Figure~\ref{fig:polymonitor_mas}). Instead of disconnected tables, each market is reconstructed as an evolving object---definition, trading, liquidity, resolution, history---so price paths, displayed depth, fills, oracle outcomes, and related external events can be inspected, compared, and traced with timing and provenance intact. Topic panels (lifecycle, funding signals, news and multimedia, macroeconomics and CPI, and sector views from crypto to sports) surface decision-relevant views: unusual probability moves, concentrated flow, liquidity shifts, cross-contract inconsistencies, and nearing resolutions. The platform thus extends the dataset from a passive archive into an inspectable instrument for hypothesis generation and reproducible lifecycle analysis.

\subsection{Multi-Agent System Panel: Evidence Integration and Decision-Oriented Analysis}
\label{app:mas_panel}

The Multi-Agent System (MAS) panel converts this heterogeneous evidence into structured, decision-oriented output (Figure~\ref{fig:polymonitor_mas}). Deterministic stages first assemble a shared evidence state---evidence, relations, quantitative checks, and memory of previous runs---from contract metadata and resolution rules, prices, book conditions, executions, flow concentration, related-market structure, oracle activity, external news, and data-quality flags. Specialized agents then examine complementary slices of this state (microstructure, catalysts, resolution conditions, calibration, cross-market relations), with a skeptic agent stress-testing conclusions before a writer agent compiles them into overview, radar, and trend views: market briefs, evidence-backed findings, themes, risks, confidence assessments, and watchlists. Each finding states what deserves attention, why, on which observations, and what future evidence would confirm or refute it, and remains linked to identifiable markets and their quantitative record; skeptic review ranks conclusions by evidence quality. The panel thus forms an auditable layer from heterogeneous observations to prioritized investigation.



\section{Conclusion}

We introduce a continuously synchronized, full-lifecycle dataset for decentralized prediction markets that unifies fragmented market metadata, on-chain executions, and oracle-resolution records into a cohesive research resource. As of the September 1, 2026 snapshot, the dataset comprises 3.29 million consolidated market identities, 1.90 billion deduplicated OrderFilled events, and 21.03 million oracle lifecycle events, with over 2.4 million identities containing both observed trading activity and verified final-resolution evidence to enable end-to-end analysis. By preserving cross-source identifiers, transaction-level provenance, and resolution histories, this infrastructure eliminates the need for ad-hoc data reconstruction and directly supports diverse downstream applications, as demonstrated through NBA probability calibration, market-implied CPI expectation reconstruction, and resolution prediction from partial execution histories. These use cases validate the dataset's versatility for rigorous economic analysis, trading research, and temporal machine learning. To facilitate community adoption, the dataset is publicly available at https://anonymous.4open.science/r/polymonitor-DBC2, accompanied by the interactive visualization dashboard Polymonitor, collectively providing a reproducible empirical foundation for studying forecasting, market microstructure, information aggregation, and oracle mechanisms, while continuous synchronization ensures long-term alignment with the evolving prediction-market ecosystem.

\newpage

\bibliography{sample-base}
\bibliographystyle{iclr2027_conference}

\newpage
\appendix
\section{ Code and Data Availability}
Once this paper is accepted, all code and data will be made publicly available, including data acquisition scripts, datasets, specific experimental scripts for the applications and results.

\section{Background and Related Work}

\subsection{Polymarket as a Premier Prediction Market}

Prediction markets trade event-contingent claims, typically binary contracts with payoff $X=\mathbf{1}\{E\}\in\{0,1\}$, where a YES price $p_t\in[0,1]$ is commonly interpreted as the market-implied probability $p_t\approx \Pr_t(E)$. Because such contracts can cover elections, sports, macroeconomic releases, geopolitical events, and many other future outcomes, their prices provide broad, continuously updated signals of collective beliefs. Polymarket implements this mechanism through YES/NO outcome tokens traded on an off-chain CLOB, with token custody, settlement, and oracle-based resolution recorded on chain. As a result, each market naturally produces three linked lifecycle layers: market metadata defining the event, OrderFilled records capturing fill-level trading and price formation, and UMA oracle records determining final resolution and payoff.

\subsection{On-chain Blockchain Datasets}

%

\begin{table}[!t]
\centering
\scriptsize
\setlength{\tabcolsep}{3.2pt}
\renewcommand{\arraystretch}{1.05}
\caption{Comparison with prior prediction-market studies and open-source on-chain financial datasets.
PM: prediction market; Meta.: market metadata; Trade: fill-level trading records; Oracle: oracle-side resolution records; Settle.: settlement information; Link: cross-layer lifecycle linking; Live: continuously updated dataset; Emp.: empirical analysis.}
\label{tab:pm_dataset_compare}

\resizebox{\textwidth}{!}{%
\begin{tabular}{@{}llccccccc l@{}}
\toprule
\textbf{Work / Dataset}
& \textbf{Focus}
& \multicolumn{4}{c}{\textbf{Lifecycle Coverage}}
& \multicolumn{3}{c}{\textbf{Properties}}
& \textbf{Scale} \tabularnewline
\cmidrule(lr){3-6}
\cmidrule(lr){7-9}
&
& \textbf{Meta.}
& \textbf{Trade}
& \textbf{Oracle}
& \textbf{Settle.}
& \textbf{Link}
& \textbf{Live}
& \textbf{Emp.}
& \tabularnewline
\midrule

\multicolumn{10}{@{}l}{\textit{Previous prediction-market studies}} \tabularnewline
\midrule
\mbox{Diercks et al.~\cite{diercks2026kalshi}}
& PM
& \cmark
& \cmark
& \xmark
& \xmark
& \xmark
& \xmark
& \cmark
& \mbox{Hundreds of Kalshi macro markets} \tabularnewline

\mbox{Ng et al.~\cite{ng2026price}}
& PM
& \cmark
& \cmark
& \xmark
& \xmark
& \xmark
& \xmark
& \cmark
& \mbox{Four platforms before the 2024 U.S. election} \tabularnewline

\mbox{Cong et al.~\cite{cong2025financial}}
& Oracle
& \xmark
& \xmark
& \cmark
& \cmark
& \xmark
& \xmark
& \cmark
& \mbox{Ethereum DeFi protocols with oracle integration} \tabularnewline
\midrule

\multicolumn{10}{@{}l}{\textit{Open-source on-chain financial datasets}} \tabularnewline
\midrule
\mbox{EX-Graph~\cite{wang2023ex}}
& Txn.
& \xmark
& \xmark
& \xmark
& \xmark
& \xmark
& \xmark
& \cmark
& \mbox{268.3M txs, 18.6M addresses, 24.3K tokens} \tabularnewline

\mbox{NFT1000~\cite{wang2024nft1000}}
& NFT
& \xmark
& \xmark
& \xmark
& \xmark
& \xmark
& \xmark
& \cmark
& \mbox{7.56M image-text pairs from 1K collections} \tabularnewline

\mbox{Cernera et al.~\cite{cernera2023token}}
& Rug pull
& \xmark
& \xmark
& \xmark
& \xmark
& \xmark
& \xmark
& \cmark
& \mbox{4.53M Ethereum tokens and 3.09M BNB tokens} \tabularnewline

\mbox{Midsummer~\cite{mongardini2025midsummer}}
& Meme
& \xmark
& \xmark
& \xmark
& \xmark
& \xmark
& \xmark
& \cmark
& \mbox{34,988 tokens across four chains} \tabularnewline
\midrule

\textbf{Ours}
& \textbf{PM}
& \textbf{\cmark}
& \textbf{\cmark}
& \textbf{\cmark}
& \textbf{\cmark}
& \textbf{\cmark}
& \textbf{\cmark}
& \textbf{\cmark}
& \mbox{\textbf{3.29M identities / 1.90B fills / 21.03M oracle events}} \tabularnewline
\bottomrule
\end{tabular}%
}
\end{table}

Data-centric research on blockchain systems has produced datasets for on-chain activity, financial behavior, and protocol risks, including transaction traces and token-flow modeling across UTXO and account-based systems \cite{gramoli2023diablo, luo2024multi, hu2024zipzap}, Ethereum account interactions, execution behaviors, and graph-structured fund movements \cite{wang2023ex, zhou2025graph}, as well as domain-specific resources for yield farming and DeFi \cite{ni2024money, zhou2023sok}, meme coins \cite{mongardini2025midsummer}, stablecoins \cite{guan2025security}, cross-chain protocols \cite{cao2026price}, NFT ecosystems \cite{wang2024nft1000}, and token-level misconduct such as rug pulls, malicious launches, and forensic analysis \cite{cernera2023token, zhou2024stop, sun2025sok, guan2024characterizing}. While these datasets have enabled large-scale blockchain analysis, they primarily focus on transfers, tokens, protocols, or risk events, rather than prediction markets as end-to-end lifecycle systems.

Prediction markets have likewise been studied as empirical systems for belief aggregation and information discovery \cite{rahman2025sok}. Prior work uses prediction-market prices as high-frequency signals for macroeconomic expectations, elections, and political or policy uncertainty \cite{diercks2026kalshi, cutting2025betting, eichengreen2025under}, and analyzes arbitrage, semantic relationships, cross-platform price discovery, and trading behavior \cite{ng2026price, saguillo2025unravelling, capponi2025semantic}. Related studies on on-chain derivatives and decentralized markets further emphasize oracle design, incentives, and cross-network information integration \cite{cong2023onchain, cong2025financial, cong2025primer, chen2021brief}. However, these works typically examine specific mechanisms or applications rather than providing a unified, continuously maintained data infrastructure. Our dataset addresses this gap by reconstructing Polymarket markets as full lifecycles linking metadata, token registration, fill-level trading, oracle interaction, disputes, resolution, and settlement. Table~\ref{tab:pm_dataset_compare} summarizes the contribution of this work.

\section{Dataset}

%

\subsection{Metadata Schema and Record Examples}
\label{app:metadata_schema}

Tables~\ref{tab:market_metadata}--\ref{tab:oracle_metadata}
describe the stored fields of the Market, OrderFilled, and Oracle
data units. Market examples come from the Trump presidential-election
contract (local market ID 422988); OrderFilled examples come
from one execution involving its YES token. Oracle examples use
explicitly identified resolution events, since stage-specific fields
are not all populated in one row. All examples were retrieved from
the databases. Long identifiers and text are abbreviated with
...; timestamps are displayed in UTC. NULL denotes
an actual SQL null, and \{\} denotes an actual empty JSON object.

\begingroup
\footnotesize
\setlength{\tabcolsep}{4pt}
\renewcommand{\arraystretch}{1.12}

\begin{longtable}{@{}
  >{\raggedright\arraybackslash}p{0.29\textwidth}
  >{\raggedright\arraybackslash}p{0.36\textwidth}
  >{\raggedright\arraybackslash}p{0.30\textwidth}@{}}
\caption{Market fields and examples from core.markets,
record id = 422988. Each row in the source table describes
one prediction contract.}\label{tab:market_metadata}\\
\toprule
Field & Description & Example \\
\midrule
\endfirsthead
\multicolumn{3}{l}{Table \thetable\ continued}\\
\toprule
Field & Description & Example \\
\midrule
\endhead
\midrule
\multicolumn{3}{r}{Continued on next page}\\
\endfoot
\bottomrule
\endlastfoot

id
& Local market identifier.
& 422988 \\

gamma\_market\_id
& Source identifier supplied by Gamma.
& 253591 \\

identity\_kind
& Source or classification of the market identity.
& canonical\_gamma \\

identity\_proof
& Structured supporting identity evidence, when populated.
& \{\} \\

event\_id
& Identifier of the parent event.
& 903193 \\

event\_slug
& URL identifier of the parent event.
& presidential-election-2024 \\

event\_title
& Display title of the parent event.
& Presidential Election Winner 2024 \\

slug
& URL identifier of the market.
& will-donald-trump-win-2024-us-presidential \\

title
& Prediction question defining the contract.
& Will Donald Trump win the 2024 US Presidential Election? \\

description
& Resolution rules, information sources, and exceptional conditions.
& This market will resolve to ``Yes'' if Donald J. Trump wins the
2024 US Presidential Election. Otherwise, this market will resolve
to ``No.'' ... \\

condition\_id
& Associated Conditional Tokens condition identifier.
& 0xdd22472e...d110917 \\

question\_id
& Associated protocol question identifier.
& 0xe3b1bc38...cec030f00 \\

oracle
& Oracle address recorded in the market metadata.
& 0x2F5e3684...c2c0aA9d \\

yes\_token\_id
& YES outcome-token identifier, represented as a decimal string.
& 2174263314...938836455 \\

no\_token\_id
& NO outcome-token identifier, represented as a decimal string.
& 4833104333...308965732 \\

clob\_token\_ids
& List of associated CLOB outcome-token identifiers.
& ["2174263314...",\newline"4833104333..."] \\

market\_type
& Contract structure: binary, categorical, or scalar.
& binary \\

enable\_neg\_risk
& Whether the market uses the negative-risk mechanism.
& true \\

category
& Source-supplied thematic category.
& politics \\

tags
& List of source-supplied classification tags.
& ["all", "potusbanner",\newline"trump", "politics", ...] \\

active
& Source-reported activity flag.
& true \\

closed
& Source-reported closure flag.
& true \\

accepting\_orders
& Whether the source reports that orders are accepted.
& false \\

enable\_order\_book
& Whether order-book functionality is enabled.
& true \\

created\_at
& Source-reported market creation timestamp.
& 2024-01-04\newline17:33:51.332 UTC \\

raw\_created\_at
& Original creation-timestamp string.
& 2024-01-04T17:33:51.332Z \\

end\_date
& Reported market end time, distinct from oracle settlement time.
& 2024-11-05\newline12:00:00 UTC \\

raw\_end\_date
& Original end-timestamp string.
& 2024-11-05T12:00:00Z \\

gamma\_updated\_at
& Source-reported Gamma update timestamp.
& 2024-11-07\newline15:23:00.928768 UTC \\

market\_state\_observed\_at
& Time at which the stored market state was observed.
& 2026-09-06\newline10:53:06.970623 UTC \\

migrated\_at
& Database bookkeeping timestamp.
& 2026-09-06\newline10:53:06.970623 UTC \\
\end{longtable}

\noindent
The status flags above are reproduced independently as stored:
active = true and closed = true coexist in this
record. The activity flag alone does not establish that orders
can currently be submitted. These metadata values are the retrieved
record's state, not a reconstruction of what was observable at
the 2024 execution time.

\begin{longtable}{@{}
  >{\raggedright\arraybackslash}p{0.29\textwidth}
  >{\raggedright\arraybackslash}p{0.36\textwidth}
  >{\raggedright\arraybackslash}p{0.30\textwidth}@{}}
\caption{OrderFilled fields from the archived
orderfilled\_fact table. All examples belong to the same
execution, identified by transaction 71230a12...27f32a3e
and log index 137.}\label{tab:orderfilled_metadata}\\
\toprule
Field & Description & Example \\
\midrule
\endfirsthead
\multicolumn{3}{l}{Table \thetable\ continued}\\
\toprule
Field & Description & Example \\
\midrule
\endhead
\midrule
\multicolumn{3}{r}{Continued on next page}\\
\endfoot
\bottomrule
\endlastfoot

tx\_hash
& Transaction hash containing the execution event.
& 71230a12...27f32a3e \\

log\_index
& Log index locating the event within its transaction.
& 137 \\

block\_number
& Block containing the execution.
& 62744628 \\

contract
& Exchange identifier; this historical record stores an exchange name.
& NegRisk CTF Exchange \\

order\_hash
& Identifier of the filled order.
& f9421bf1...21dcfbf4 \\

market\_id
& Local identifier linking the execution to its market.
& 422988 \\

condition\_id
& Condition associated with the traded asset.
& 0xdd22472e...d110917 \\

token\_id
& Outcome-token identifier, represented as a hexadecimal string.
& 3011e4ed...6e3a43e7 \\

outcome\_code
& Binary outcome encoding: 1 for YES and 2 for NO.
& 1 (YES) \\

maker
& Maker address retained from the execution record.
& 5bffcf56...84d6ffbe \\

taker
& Taker address retained from the execution record.
& c5d563a3...5220f80a \\

side\_code
& Stored trade-side encoding: 1 for BUY and 2 for SELL.
& 1 (BUY) \\

price
& Execution price in collateral units per outcome-token share.
& 0.5100000000 \\

size
& Executed quantity in outcome-token shares.
& 242388.9200000000 \\

maker\_amount
& Raw integer amount of the maker-side asset.
& 242388920000 \\

taker\_amount
& Raw integer amount of the taker-side asset.
& 123618349200 \\

fee
& Event-reported fee in raw contract units, excluding gas costs.
& 0 \\

ingested\_at
& Timestamp stored for analytical ingestion.
& 2026-06-02 04:05:25 \\
\end{longtable}

\noindent
The Market decimal token identifier and the OrderFilled hexadecimal
token identifier encode the same integer. Hashes and participant
addresses in this archived execution are stored without the
0x prefix. The table reproduces the historical side code;
participant-level direction must follow the relevant source convention
rather than assuming that every acquisition version uses the same
maker/taker perspective. The ClickHouse ingested\_at example
is reproduced as returned by the database, without adding a timezone
suffix to its stored display representation.

\begin{longtable}{@{}
  >{\raggedright\arraybackslash}p{0.29\textwidth}
  >{\raggedright\arraybackslash}p{0.36\textwidth}
  >{\raggedright\arraybackslash}p{0.30\textwidth}@{}}
\caption{Oracle fields from oracle.oracle\_events.
Examples use settlement record S (id = 141207) unless
marked R (request), P (proposal), C (Conditional Tokens settlement),
or D (a dispute in another market).}\label{tab:oracle_metadata}\\
\toprule
Field & Description & Example \\
\midrule
\endfirsthead
\multicolumn{3}{l}{Table \thetable\ continued}\\
\toprule
Field & Description & Example \\
\midrule
\endhead
\midrule
\multicolumn{3}{r}{Continued on next page}\\
\endfoot
\bottomrule
\endlastfoot

id
& Local identifier of the oracle event.
& 141207 \\

tx\_hash
& Transaction hash containing the oracle event.
& 0xa31de7f7...4ec1515f \\

log\_index
& Log index locating the event within its transaction.
& 155 \\

block\_number
& Block containing the oracle event.
& 63956323 \\

event\_time
& Blockchain timestamp associated with the event.
& 2024-11-06\newline13:32:47 UTC \\

raw\_event\_time
& Original timestamp representation, when retained.
& NULL (actual stored value) \\

event\_status
& Resolution stage: request, propose, dispute, or settle.
& settle \\

external\_market\_id
& External market identifier available from the source or mapping.
& 253591 [C] \\

market\_id
& Local identifier of the associated market.
& 422988 \\

market\_title
& Title of the associated prediction contract.
& Will Donald Trump win the 2024 US Presidential Election? \\

source\_adapter
& Associated adapter or requesting contract, depending on the source.
& 0x2f5e3684...c2c0aa9d \\

source\_oracle
& Emitting oracle or resolution contract.
& 0xee3afe34...af887c24 \\

adapter\_question\_id
& Question identifier used in adapter linkage.
& 0xc2d6714f...79da1605 \\

question\_id
& Associated protocol question identifier.
& 0xe3b1bc38...cec030f00 \\

condition\_id
& Associated Conditional Tokens condition identifier.
& 0xdd22472e...d110917 \\

matched\_by
& Method recorded for the event-to-market association.
& question\_id \\

string\_raw
& Decoded ancillary text or serialized resolution payload.
& q: title: Will Donald Trump win the 2024 US Presidential Election?, description: ... \\

p1
& First parsed resolution alternative; No for this question.
& 0 \\

p2
& Second parsed resolution alternative; Yes for this question.
& 1 \\

proposed\_price
& Normalized UMA proposal value, not a trading price.
& 1 [P] \\

settled\_price
& Normalized UMA settlement value, interpreted using the resolution rules.
& 1 (YES) \\

payout
& UMA event payout in raw units, or a Conditional Tokens payout-numerator vector.
& 100500000000 [S];\newline[1, 0] [C] \\

settlement\_recipient
& Stored recipient-related enrichment; the historical extraction derives this field from proposer.
& 0xcc400c09...6223e37d \\

requester
& Address requesting oracle resolution.
& 0x2f5e3684...c2c0aa9d \\

proposer
& Address proposing the resolution value.
& 0xcc400c09...6223e37d \\

disputer
& Address challenging a proposed resolution.
& 0xd7e93d50...6ff6e90 [D] \\

request\_transaction
& Reference to the request-stage transaction, when populated.
& 0xe687e360...00fbb8e4 [R] \\

proposal\_transaction
& Reference to the proposal-stage transaction, when populated.
& 0x241c3d03...6bc4ca55 \\

settlement\_transaction
& Reference to the settlement-stage transaction, when populated.
& 0xa31de7f7...4ec1515f \\

created\_at
& Database insertion timestamp, distinct from event time.
& 2026-03-09\newline08:23:08 UTC \\
\end{longtable}

\noindent
Oracle example provenance is as follows: R is request record
636002; P is proposal record 601368; S is UMA
settlement record 141207; and C is Conditional Tokens
settlement record 11820855. All four reference market
422988. D is dispute record 1786175, associated
with market 253526, transaction
0x427bff76...ebd24403, log index 1923.
It illustrates a populated disputer field and does not imply a
dispute in the Trump example. The actual disputer value in S is
the zero address. Likewise, S stores an empty string in
proposed\_price; the nonempty example is taken from P.
The populated historical settlement\_recipient field
is not independent evidence of a verified recipient transfer.

\endgroup

\begin{table}[t]
\centering
\caption{
Data domains and records comprising the lifecycle dataset.
}
\label{tab:pipeline_services}
\small
\begin{tabular}{p{0.22\linewidth} p{0.32\linewidth} p{0.40\linewidth}}
\toprule
Domain & Primary source & Collected records \\
\midrule
Market metadata
& Polymarket Gamma and CLOB APIs
& Market, event, condition, and outcome-token metadata \\

Executions
& Polygon transaction logs
& Raw and normalized \texttt{OrderFilled} events with block and transaction
  provenance \\

Oracle
& Polygon logs (UMA, CTF, resolution modules)
& Proposal, dispute, settlement, and market-resolution records \\
\bottomrule
\end{tabular}
\end{table}

\subsection{Data Collection Pipeline}
\label{sec:appendix_pipeline}

The dataset is assembled through a fault-tolerant pipeline that continuously
collects three data domains from Polymarket and the Polygon blockchain: market
metadata, on-chain executions, and oracle resolution events. Each domain is
collected by an independent worker maintaining a persistent progress
checkpoint. To tolerate delayed API responses, transient source outages, and
short chain reorganizations, workers resume from a bounded overlap preceding
the latest confirmed checkpoint rather than from the most recent observation.
Records are deduplicated via stable source identifiers, and checkpoints
advance only after the corresponding interval passes validation. This
checkpoint-and-replay design ensures that interrupted collections resume
deterministically without duplicating existing observations or silently
dropping unresolved intervals. Table~\ref{tab:pipeline_services} summarizes
the data sources and record types.

\paragraph{Unified synchronization protocol.}
All three collectors implement the same synchronization logic: initiate from
the latest validated checkpoint, replay a bounded overlap with the preceding
interval, persist newly observed records under stable identifiers, and advance
the checkpoint upon successful interval validation. Blockchain queries are
restricted to a confirmed chain head with a 20-block confirmation margin;
large log ranges are partitioned into smaller block intervals, and logs
explicitly marked as removed due to reorganizations are discarded. Incomplete
or failed intervals are flagged for retry rather than marked as complete.
API-based sources follow an analogous protocol, periodically revisiting recent
records to capture delayed metadata updates.

\paragraph{Market metadata.}
Market metadata are extracted from the Polymarket Gamma and CLOB interfaces,
querying both market- and event-level endpoints for active and closed
contracts. Pagination relies on monotonic source cursors, supplemented by a
bounded time-window overlap (two hours in production) to recover records that
become visible after the initial pass.

Records are normalized around the Polymarket condition identifier, which
serves as the primary key linking market descriptions, event structures, and
outcome tokens. Original platform identifiers, question IDs, market statuses,
and token labels are preserved alongside this canonical representation.
Metadata updates modify existing records in place rather than creating
duplicate logical markets. A periodic full historical sweep, independent of
the continuous collector, verifies that incremental synchronization does not
systematically omit older records.

Because metadata APIs and on-chain activity are not perfectly synchronized,
market registration does not strictly precede trading activity. A token that
first appears in an execution record before its market metadata are available
is retained and linked retrospectively once the corresponding market
registration is recovered.

\paragraph{On-chain executions.}
Trading activity is reconstructed from Polygon logs emitted by Polymarket
exchange contracts. The collector scans all supported exchange deployments,
retaining both the raw blockchain event and a normalized execution record.
Observations are uniquely identified by the tuple \texttt{(contract address,
transaction hash, log index)}, ensuring that repeated scans of overlapping
block intervals remain strictly deterministic.

Block timestamps are fetched in batches and attached to each event prior to
normalization. The execution record is then mapped to its corresponding
condition and outcome token via the market registry, extracting the canonical
market ID, traded outcome, direction, price, and quantity while preserving
transaction-level provenance. Full transaction receipts are archived to enable
independent verification. Executions referencing tokens with unavailable
metadata undergo deferred resolution: the event is retained and reprocessed
once the mapping becomes available, preventing temporary metadata lag from
creating permanent gaps in the trading record.

\paragraph{Oracle and resolution records.}
Resolution activity is reconstructed from Polygon logs associated with the UMA
oracle infrastructure, Conditional Token Framework (CTF) contracts,
negative-risk adapters, and newer resolution modules. This collector follows
the unified synchronization protocol with an additional short rewind upon
resuming from a checkpoint to capture events affected by recent
reorganizations.

Rather than recording only the final market outcome, we preserve the full
sequence of resolution actions. For UMA-based markets this includes request,
proposal, dispute, and settlement events; for CTF-based modules it includes
preparation and settlement events. Ancillary-data hashes, request IDs,
question IDs, and condition IDs are retained to facilitate linking oracle
activity back to specific markets. Oracle records lacking immediate metadata
mappings are flagged for deferred reconciliation rather than assigned via
heuristics.

\paragraph{Reproducibility and data integrity.}
The unified checkpoint-and-replay protocol strictly separates source
incompleteness from the actual absence of an event. Overlapping scans
reproduce existing observations deterministically, while newly recovered
metadata or blockchain events extend the existing lifecycle history without
altering prior records. As a result, market metadata, executions, and oracle
events can be updated continuously while preserving their original provenance,
temporal ordering, and evidential integrity.

\subsection{Detailed Data Statistics and Coverage}
\label{data_statistics}

\paragraph{Snapshot definition.}
We report dataset statistics using a fixed event-time cutoff of
September 1, 2026, 00:00 UTC, excluding records at or after the cutoff.
For Polygon-derived records, the last included block is
\textbf{93,010,514}. All statistics below are computed under this common
boundary so that the market, execution, and resolution layers can be
compared consistently. As summarized in Table~\ref{tab:dataset_scale},
the snapshot contains \textbf{3,290,907} consolidated market and
protocol-condition identities, \textbf{1,901,194,769} deduplicated
{OrderFilled} events, and \textbf{21,028,317} oracle-, adapter-,
and protocol-module lifecycle events. Historical FPMM/AMM trading is
retained separately, contributing an additional \textbf{2,052,409}
events from earlier Polymarket trading mechanisms.

\begin{table}[t]
\centering
\small
\caption{Scale of the fixed Polymarket lifecycle snapshot. The three
primary layers use different units of observation and are linked through
market, condition, question, and outcome-token identifiers.}
\label{tab:dataset_scale}
\begin{tabular}{lrr}
\toprule
\textbf{Dataset layer} & \textbf{Unit} & \textbf{Count} \\
\midrule
Market catalog
    & Market / protocol condition
    & 3,290,907 \\
OrderFilled
    & Deduplicated fill event
    & 1,901,194,769 \\
Oracle and resolution
    & Lifecycle event
    & 21,028,317 \\
Historical FPMM/AMM
    & Trading event
    & 2,052,409 \\
\bottomrule
\end{tabular}
\end{table}

\paragraph{Market catalog composition.}
The market layer is constructed as an identity catalog rather than as a
direct count of rows returned by a single API. It consolidates official
market identities with historical and protocol-level conditions that are
required to connect earlier or structurally different Polymarket
deployments to their trading and resolution records.
Table~\ref{tab:market_identity_composition} shows the resulting
composition. Official Gamma markets constitute the large majority,
with \textbf{2,939,507 identities (89.32\%)}. Historical combinatorial
conditions contribute \textbf{179,092 (5.44\%)}, while
protocol-structural conditions contribute \textbf{171,808 (5.22\%)}.
The remaining identities are obtained through official combinatorial,
CLOB-derived, and chain-derived reconstruction paths. This organization
allows the dataset to cover multiple generations of Polymarket while
maintaining a common identity layer for downstream joins.

\begin{table}[t]
\centering
\small
\caption{Composition of the consolidated market catalog.}
\label{tab:market_identity_composition}
\begin{tabular}{lrr}
\toprule
\textbf{Identity type} & \textbf{Count} & \textbf{Share} \\
\midrule
Official Gamma market
    & 2,939,507 & 89.32\% \\
Historical combinatorial condition
    & 179,092 & 5.44\% \\
Protocol-structural condition
    & 171,808 & 5.22\% \\
Canonical CLOB identity
    & 481 & 0.015\% \\
Official combinatorial identity
    & 5 & $<$0.001\% \\
Chain-derived identity
    & 14 & $<$0.001\% \\
\midrule
\textbf{Total}
    & \textbf{3,290,907} & \textbf{100\%} \\
\bottomrule
\end{tabular}
\end{table}

The consolidated count also prevents source-specific bookkeeping records
from artificially inflating the number of markets. A direct
{created\_at}-based count of the underlying market table yields
3,500,538 rows. Identity reconstruction additionally recovers
\textbf{297,826} valid identities from creation or protocol-preparation
evidence and removes \textbf{507,457} historical placeholder rows,
resulting in
\[
3,500,538 + 297,826 - 507,457
= 3,290,907.
\]
Of the removed placeholders, \textbf{507,411} have established
consolidation mappings. We therefore use 3,290,907 as the catalog
denominator throughout the paper rather than treating every observed
source row or protocol condition as an independent market.

\paragraph{Execution coverage.}
The execution layer contains \textbf{1,901,194,769} unique
{OrderFilled} logs after event-level deduplication. The analytical
fact table accounts for \textbf{1,901,194,648} of these events, while an
additional \textbf{121} raw events absent from that projection were
individually verified against their on-chain transaction receipts.
During reconciliation, \textbf{2,002} duplicated raw copies were
identified and excluded. Among all deduplicated fills,
\textbf{1,901,194,698} can be linked directly to identities in the
reported market catalog; the remaining \textbf{71} fills correspond to
only \textbf{21} genuine combinatorial conditions outside the catalog
snapshot. Thus, essentially the entire observed OrderFilled stream has
an identified market or protocol-condition association.

At the catalog level, \textbf{2,441,701 identities (74.20\%)} have at
least one observed OrderFilled execution before the cutoff. Since one
market can generate many fills, the dataset simultaneously provides
broad market coverage and extremely dense transaction-level histories.
The historical trading layer further contains \textbf{2,052,409}
FPMM/AMM events; including this earlier mechanism identifies an
additional \textbf{4,771} catalog identities with historical trading
activity but no OrderFilled record.

\paragraph{Oracle and resolution composition.}
The resolution layer contains \textbf{21,028,317} events and records
considerably more than final outcome labels. As shown in
Table~\ref{tab:oracle_composition}, it reconstructs intermediate
lifecycle activity across CTF, UMA, adapters, combinatorial contracts,
and newer protocol modules. The largest components include
\textbf{3,338,899} CTF condition-preparation events,
\textbf{3,025,824} CTF final-result writes,
\textbf{2,340,129} UMA requests,
\textbf{2,123,464} proposals,
\textbf{2,122,993} UMA settlements,
\textbf{2,333,185} adapter initializations, and
\textbf{2,110,388} {QuestionResolved} events. The dataset also
contains \textbf{4,334} observed UMA disputes, preserving the less common
contested-resolution path rather than retaining only final labels.

\begin{table}[t]
\centering
\small
\caption{Composition of the oracle and protocol-resolution layer.
Intermediate lifecycle events are preserved in addition to final
resolution records.}
\label{tab:oracle_composition}
\begin{tabular}{llr}
\toprule
\textbf{Source} & \textbf{Event type} & \textbf{Count} \\
\midrule
CTF & Condition preparation
    & 3,338,899 \\
CTF & Final result
    & 3,025,824 \\
UMA & Request
    & 2,340,129 \\
UMA & Proposal
    & 2,123,464 \\
UMA & Dispute
    & 4,334 \\
UMA & Settlement / ruling
    & 2,122,993 \\
Adapter & Initialization
    & 2,333,185 \\
Adapter & Reset
    & 3,927 \\
Adapter & {QuestionResolved}
    & 2,110,388 \\
Adapter & {QuestionSettled}
    & 2,932 \\
Protocol module & {MigrationConditionRegistered}
    & 842,941 \\
Protocol module & {EventPrepared}
    & 34,724 \\
Combinatorial module & {CombinatorialConditionPrepared}
    & 537,438 \\
Protocol module & {ResultReported}
    & 744,905 \\
Protocol module & {ConditionResolved}
    & 744,905 \\
Protocol module & {MigrationResolved}
    & 717,327 \\
Module upgrade & {Upgraded}
    & 2 \\
\midrule
\multicolumn{2}{l}{\textbf{Total}}
    & \textbf{21,028,317} \\
\bottomrule
\end{tabular}
\end{table}

CTF final-result writes and protocol-module
{ConditionResolved} events alone contribute
\textbf{3,770,729 direct result events}. Event counts are intentionally
kept distinct from market-level resolution counts because one market can
generate multiple lifecycle observations and some combinatorial outcomes
are derived from their constituent conditions. Under our strict
market-level resolution criterion, \textbf{3,077,506 catalog identities
(93.52\%)} have verified final-resolution evidence.

\paragraph{Cross-layer lifecycle coverage.}
The principal advantage of the dataset is that its three layers can be
connected at market level rather than analyzed as independent sources.
Table~\ref{tab:lifecycle_coverage} reports the joint coverage of
OrderFilled trading and final-resolution evidence. Of all
\textbf{3,290,907} catalog identities,
\textbf{2,402,991 (73.02\%)} contain both observed OrderFilled activity
and verified final-resolution evidence. These records form more than
\textbf{2.4 million directly linked trading-to-outcome lifecycles}, for
which contract semantics, transaction histories, and realized outcomes
can be analyzed jointly.

\begin{table}[t]
\centering
\small
\caption{Market-level coverage of trading and verified final-resolution
evidence at the fixed snapshot cutoff.}
\label{tab:lifecycle_coverage}
\begin{tabular}{lrr}
\toprule
\textbf{Observed lifecycle evidence}
    & \textbf{Count} & \textbf{Share} \\
\midrule
OrderFilled + final resolution
    & 2,402,991 & 73.02\% \\
OrderFilled observed
    & 2,441,701 & 74.20\% \\
Verified final resolution
    & 3,077,506 & 93.52\% \\
\bottomrule
Total catalog identities
    & 3,290,907 & 100\% \\
\bottomrule
\end{tabular}
\end{table}

This overlap is particularly useful for supervised and longitudinal
research because the same canonical identity connects the market
definition, the sequence of observed transactions, and the eventual
outcome. The resulting records can therefore support analyses of
probability calibration, price discovery, market microstructure,
temporal forecasting, participant behavior, and resolution-aware
machine learning without reconstructing these links separately for each
task.

\paragraph{Coverage interpretation.}
The three layers represent different stages of the prediction-market
lifecycle and therefore need not have identical market-level coverage.
To make this distinction explicit, Table~\ref{tab:coverage_partition}
partitions the complete catalog into four mutually exclusive groups.
Beyond the \textbf{2,402,991} identities containing both types of
evidence, \textbf{38,710 (1.18\%)} contain OrderFilled activity but do
not yet satisfy the strict final-resolution criterion at the snapshot
boundary, \textbf{674,515 (20.50\%)} have final-resolution evidence
without an observed OrderFilled event, and \textbf{174,691 (5.31\%)}
contain neither within the cutoff. These categories preserve the
difference between market existence, trading activity, and settlement
rather than collapsing them into a single coverage indicator.

\begin{table}[t]
\centering
\small
\caption{Mutually exclusive partition of the catalog according to
observed trading and final-resolution evidence.}
\label{tab:coverage_partition}
\begin{tabular}{lrr}
\toprule
\textbf{Coverage state} & \textbf{Count} & \textbf{Share} \\
\midrule
Trading and final resolution
    & 2,402,991 & 73.0191\% \\
Trading only
    & 38,710 & 1.1763\% \\
Final resolution only
    & 674,515 & 20.4963\% \\
Neither within snapshot boundary
    & 174,691 & 5.3083\% \\
\midrule
\textbf{Total}
    & \textbf{3,290,907} & \textbf{100\%} \\
\bottomrule
\end{tabular}
\end{table}

The unmatched categories largely reflect differences in lifecycle stage
and trading mechanism rather than a single notion of missing data.
Among the \textbf{38,710} trading-only identities, for example,
\textbf{10,343} obtain their final settlement after the snapshot cutoff
and \textbf{23,276} have scheduled end dates after the cutoff.
Among the \textbf{674,515} identities with final-resolution evidence
but no OrderFilled record, \textbf{89,021} are currently reported by
the official source with zero trading volume, while
\textbf{4,771} contain historical FPMM/AMM trading instead of
OrderFilled activity. These cases illustrate why trading coverage and
resolution coverage are reported separately.

Finally, the oracle layer also contains protocol conditions that are
useful for reconstructing the resolution system but are not automatically
counted as independent markets. In particular,
\textbf{1,428,709} additional condition identifiers appear in
oracle-preparation records but were not mapped into the consolidated
market catalog at the snapshot boundary. We retain these observations
at the protocol layer rather than inflating the market count, preserving
a clear distinction between an observed on-chain condition and a
validated market identity.

\subsection{observation and analysis}

\subsubsection{Market}
\begin{figure*}[htbp]
    \centering
    \captionsetup{font=small}

    \begin{subfigure}[t]{0.32\textwidth}
        \centering
        \includegraphics[width=\linewidth,trim=5 5 5 5,clip]{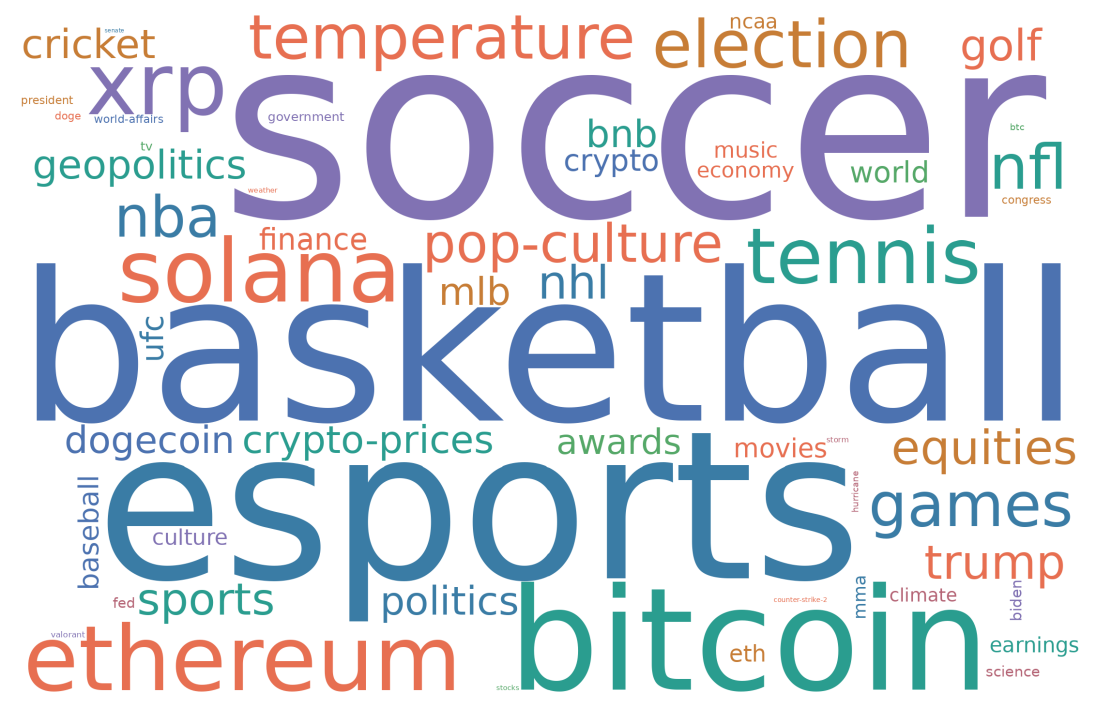}
        \caption{Primary topic keyword cloud.}
        \label{fig:market_primary_topic_keyword_cloud}
    \end{subfigure}
    \hfill
    \begin{subfigure}[t]{0.32\textwidth}
        \centering
        \includegraphics[width=\linewidth,trim=5 5 5 5,clip]{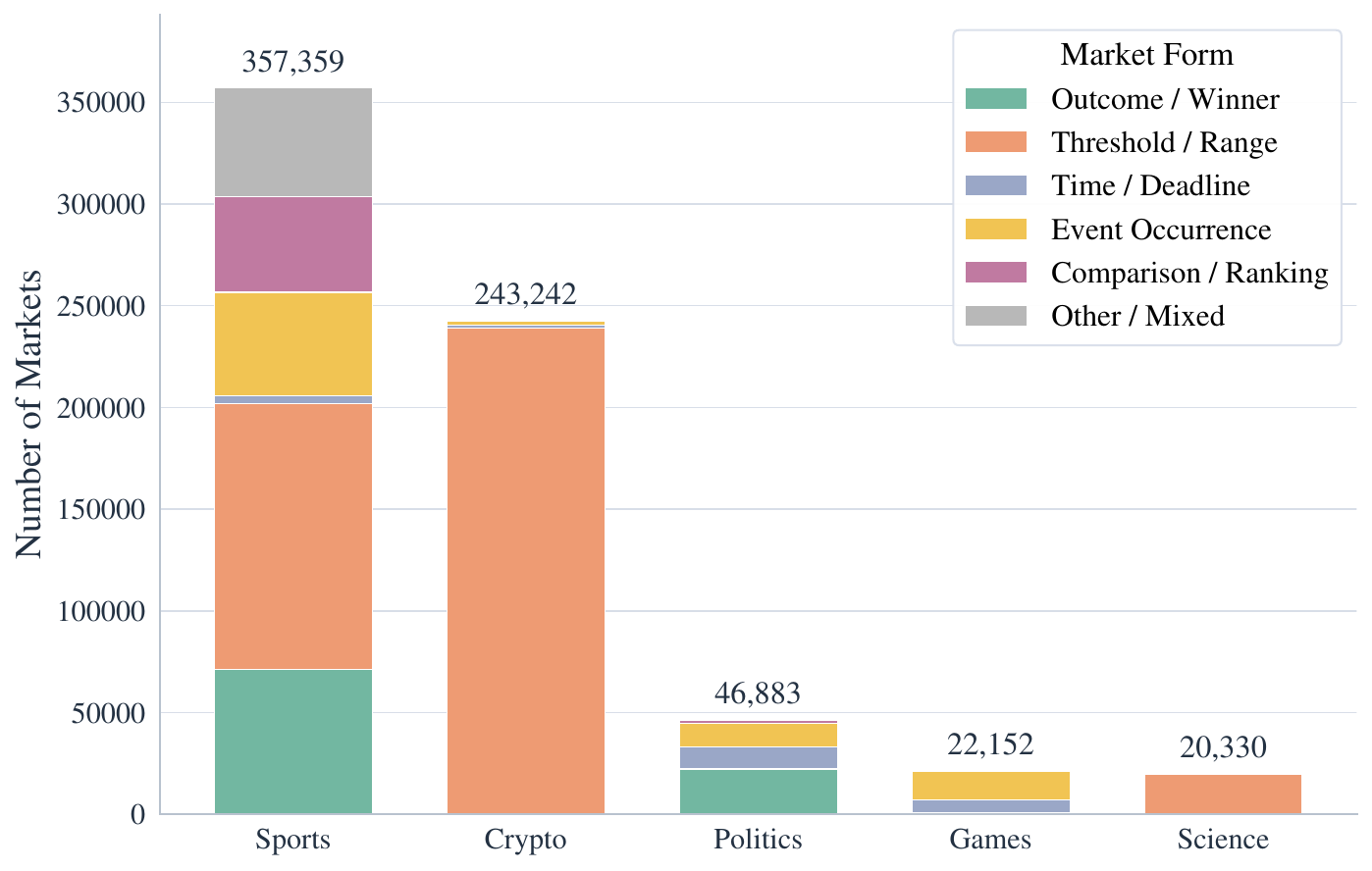}
        \caption{Top 5 primary topics by count.}
        \label{fig:market_primary_topic_top5_counts}
    \end{subfigure}
    \hfill
    \begin{subfigure}[t]{0.32\textwidth}
        \centering
        \includegraphics[width=\linewidth]{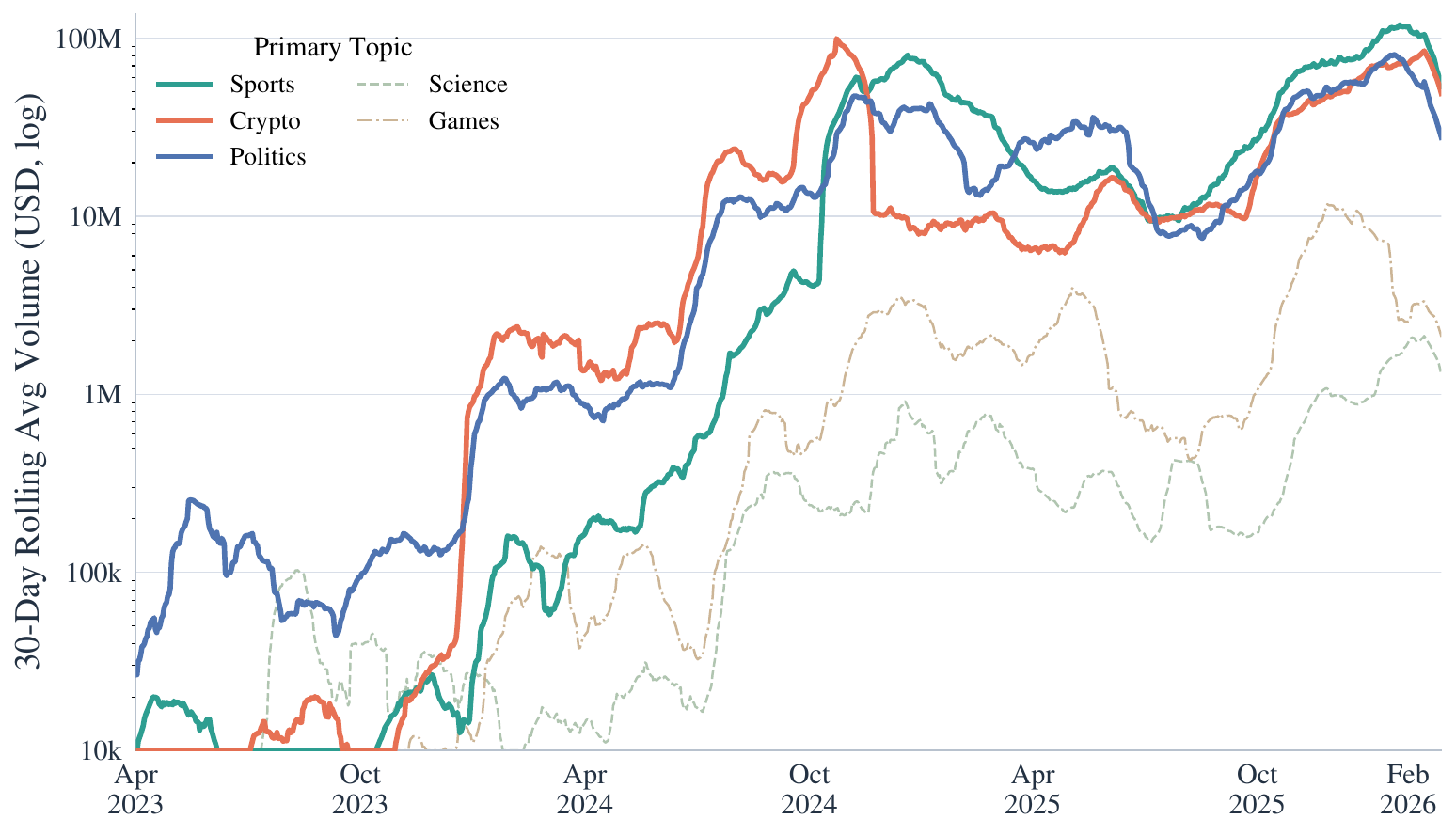}
        \caption{30-day rolling average trading volume by primary topic (log).}
        \label{fig:market_primary_topic_volume_timeseries}
    \end{subfigure}

    \caption{Primary topic analysis of Polymarket markets. (a) Keyword cloud. (b) Top 5 topics by market count. (c) 30-day rolling average trading volume (log scale).}
    \label{fig:market_primary_topic_combined}
\end{figure*}

We collected market tags from Polymarket metadata and organized them into two complementary views of topic structure. Figure \ref{fig:market_primary_topic_keyword_cloud} presents a keyword cloud of primary topic terms. The keyword cloud reveals a highly concentrated topical landscape, with sports, basketball, and soccer standing out most prominently, alongside major crypto-related terms such as bitcoin, ethereum, solana, and xrp. This pattern indicates that user attention on Polymarket is centered primarily on sports competitions and crypto-related speculation, while other themes such as politics, geopolitics, temperature, pop culture, and science remain visible but comparatively less dominant. In addition, the sports cluster is semantically rich, covering multiple leagues and game types, whereas the crypto cluster is more tightly organized around major assets and price-oriented narratives.

Figure \ref{fig:market_primary_topic_top5_counts} further quantifies this concentration. Sports is the largest category, with 357,359 markets, followed by Crypto with 243,242 markets; both far exceed Politics, Games, and Science, which contain 46,883, 22,152, and 20,330 markets, respectively. The composition across market forms is also notably different: Crypto markets are overwhelmingly dominated by threshold/range contracts, consistent with price-level and target-based speculation, whereas Sports markets are distributed across a wider variety of forms, including outcome/winner, event occurrence, comparison/ranking, and other structures. 

Figure \ref{fig:market_primary_topic_volume_timeseries} shows the 30-day rolling average trading volume by primary topic on a log scale and highlights strong temporal heterogeneity across categories. Sports, Crypto, and Politics maintain the highest trading volume over most of the observation period, while Science and Games remain at a lower level. The curves also exhibit several sharp rises, suggesting that trading activity is strongly affected by external events and topic-specific shocks rather than evolving smoothly over time. In particular, Sports and Crypto display both high baseline activity and repeated volume surges, indicating that they are not only large in market count but also dominant in realized trading activity.

\begin{figure}[htbp]
    \centering
    \begin{subfigure}[b]{0.32\linewidth}
        \centering
        \includegraphics[width=\linewidth]{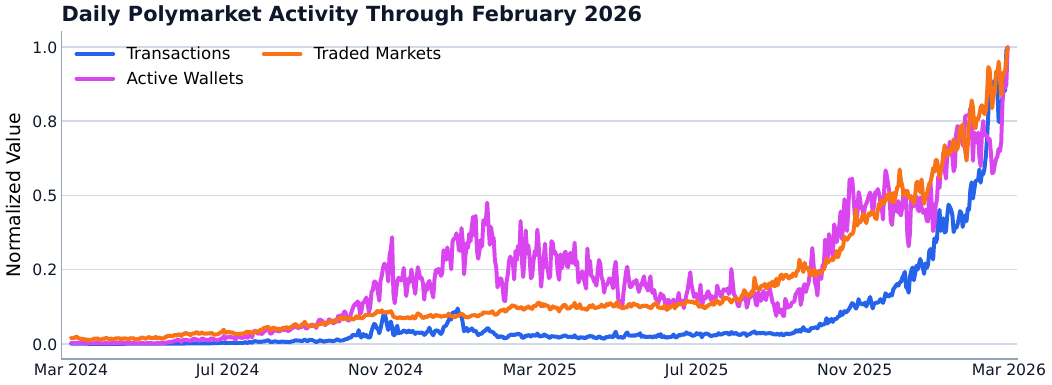}
        \caption{Normalized daily activity on Polymarket}
        \label{fig:dailyCount}
    \end{subfigure}
    \hfill
    \begin{subfigure}[b]{0.32\linewidth}
        \centering
        \includegraphics[width=\linewidth]{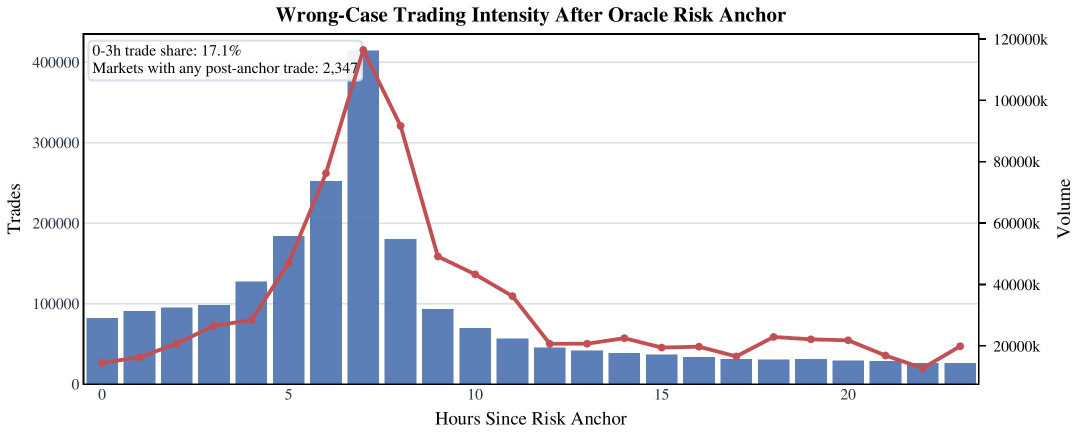}
        \caption{Wrong-case trading intensity after oracle risk anchor}
        \label{fig:wrongCaseTop}
    \end{subfigure}
    \hfill
    \begin{subfigure}[b]{0.32\linewidth}
        \centering
        \includegraphics[width=\linewidth]{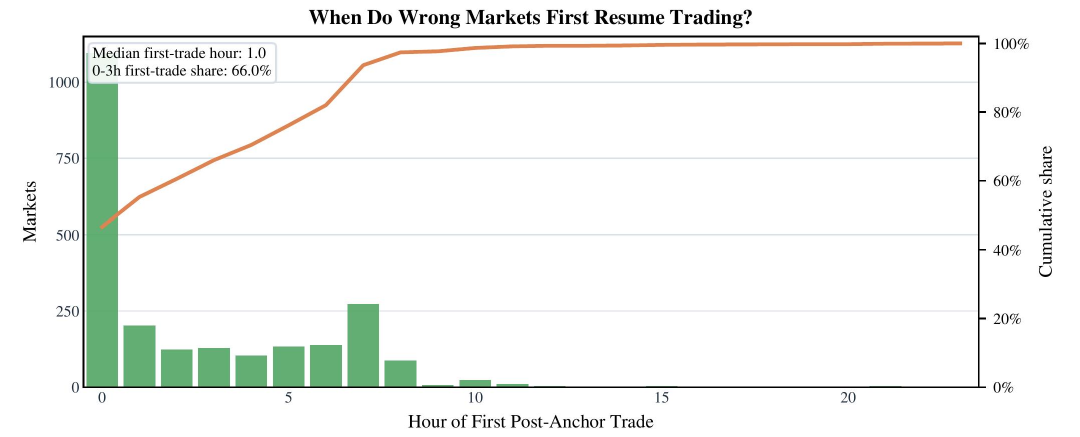}
        \caption{When do wrong markets first resume trading?}
        \label{fig:wrongCaseBottom}
    \end{subfigure}
    \caption{Analysis of Polymarket trading activity and oracle trading.}
    \label{fig:combinedAnalysis}
\end{figure}

\subsubsection{Orderfilled}

Figure \ref{fig:dailyCount} presents the normalized daily evolution of three activity indicators derived from OrderFilled events: transactions, active wallets, and traded markets. All three series stayed at low levels through the first half of 2024, indicating limited adoption in the early stage of the observation period. Starting from late 2024, the platform entered its first expansion phase. Among the three indicators, active wallets rose earlier and more sharply, reaching a clear local peak around the turn of 2024--2025. This increase was also temporally aligned with the 2024 U.S. presidential election, which likely attracted substantial attention and brought a large number of users and trades to the platform. By contrast, transactions showed only moderate increases during the same period, while traded markets expanded more steadily, implying that market participation broadened before trading depth fully caught up.

A second and stronger growth phase started in late 2025. From this point onward, traded markets and active wallets both increased rapidly, and transactions accelerated with an even steeper slope near the end of the observation window. The simultaneous rise of these three indicators suggested that Polymarket’s later expansion was not driven by a single margin, but by joint growth in user participation, market coverage, and realized trading activity. Another notable feature is that traded markets remained consistently above transactions for most of the period after normalization, which indicates that the expansion of market variety was sustained even during intervals when transaction growth was temporarily slower. Overall, the figure shows a transition from an election-related participation surge in late 2024 to a broader platform-wide scaling phase, with the latter becoming especially pronounced in the months leading to February 2026.

\subsubsection{Oracle}

We define an oracle-risk market: the market has at least one dispute event. The risk-event anchor is defined as the first dispute timestamp when a dispute exists; otherwise, it is defined as the timestamp of the last propose event. We further define continued betting as the presence of real trades within 24 hours after this risk anchor.

Under the UMA resolution path, we identify 2,358 disputed markets, which account for less than 1\% of settled UMA markets. This shows that oracle dispute events are rare overall. However, once such events occur, trading usually does not stop. Among wrong-case markets, 2,347 still record trades within 24 hours after the risk anchor, indicating that traders often continue to bet and update positions even after the market enters an oracle-risk state.

As shown in the figure \ref{fig:wrongCaseTop} and \ref{fig:wrongCaseBottom}, the top panel shows the hourly trading intensity after the oracle-risk anchor for wrong-case markets. Two patterns are clear. First, post-anchor trading starts quickly and remains active throughout the first several hours. Second, the highest trading intensity is not observed immediately at the anchor time, but several hours later, with the trade-count peak appearing around hour 7. This suggests that market participants do not react only instantaneously. Instead, a small group of traders enters early, while larger trading activity accumulates with some delay. In other words, wrong-case markets exhibit a two-stage response pattern: early speculative re-entry followed by later concentration of trading activity.

The bottom panel reports when wrong-case markets first resume trading after the oracle-risk anchor. The distribution is heavily concentrated in the earliest hours: 66.0\% of markets record their first post-anchor trade within 0--3 hours, and the median first-trade time is only 1 hour. The cumulative curve also rises steeply at the beginning and approaches saturation within the first several hours. This indicates that market reopening in practice is very fast once an oracle-risk event becomes visible.

\section{Additional Application Details}
\label{app:applications}

This section provides additional data construction, experimental settings,
and quantitative results for the three representative applications presented
in the main text. The applications deliberately use different portions of the
lifecycle dataset: the NBA study combines market metadata and pre-game
executions with realized game outcomes; the CPI study reconstructs continuous
expectations from collections of related bucket markets; and the supervised
learning study links partial execution histories to final oracle settlements.
Together, they illustrate how the same underlying lifecycle data can be
reorganized for domain-specific economic analysis and general machine-learning
tasks.

\subsection{NBA Probability Calibration}
\label{app:nba}

\paragraph{Dataset construction.}
We first identify single-game NBA winner markets from the market metadata
and associate each contract with its two competing teams, outcome tokens,
and game start time. Starting from 4,332 candidate markets, the processing
pipeline retains 4,103 cleaned markets. The resulting calibration dataset
contains 4,099 markets and 8,198 team-level observations, of which
3,901 markets (7,802 team-level observations) can be matched to realized
game outcomes. The final calibration experiment contains
1,627 training markets and 1,136 test markets. Games from seasons before
2026 form the training set, while the 2026 season is held out for testing.

\begin{table}[t]
\centering
\small
\caption{Data construction for the NBA probability-calibration application.}
\label{tab:app_nba_data}
\begin{tabular}{lr}
\toprule
\textbf{Stage} & \textbf{Count} \\
\midrule
Candidate NBA markets & 4,332 \\
Cleaned markets & 4,103 \\
Calibration markets & 4,099 \\
Calibration team-level rows & 8,198 \\
Markets with matched outcomes & 3,901 \\
Training markets & 1,627 \\
Test markets & 1,136 \\
\bottomrule
\end{tabular}
\end{table}

For every team outcome token, we retain executions satisfying
\(t_i < t_{\mathrm{start}}\), where \(t_{\mathrm{start}}\) denotes the
actual game start time. When the raw asset identifiers and integer amounts
are available, execution prices are reconstructed directly from the
collateral and outcome-token amounts; otherwise, the normalized stored price
is used. We retain executions with positive size and valid prices in
\([0,1]\), and exclude markets whose temporal or semantic alignment does not
support a single-game interpretation. We additionally require the final
observed pre-game execution to occur within six hours of tip-off.

The pre-game probability of team \(m\) is defined as the
trade-size-weighted mean of all valid pre-game executions,

\[
p_m^{\mathrm{pre}}
=
\frac{
    \sum_{i:t_i<t_{\mathrm{start}}} s_i p_i
}{
    \sum_{i:t_i<t_{\mathrm{start}}} s_i
},
\]

where \(p_i\) and \(s_i\) are the execution price and number of shares of
fill \(i\), respectively. The weighting variable is therefore executed
share quantity rather than dollar notional.

\paragraph{Outcome labels and calibration.}
Realized NBA outcomes are matched using TheSportsDB by normalized team names
and game dates, allowing at most a one-day date difference. The winning team
receives label one and the losing team label zero. We fit a one-dimensional
monotonic isotonic-regression mapping \(g(p)\) on the training markets and
compare the raw probability \(p\) with the calibrated probability \(g(p)\)
on the held-out season.

We evaluate probability quality using Brier score, Log Loss, expected
calibration error (ECE), and maximum calibration error (MCE). ECE and MCE
use 20 equally spaced probability bins. As shown in
Table~\ref{tab:app_nba_calibration}, the raw market probability performs
slightly better than the isotonic transformation on all four metrics.

\begin{table}[t]
\centering
\small
\caption{NBA calibration results on 2,272 held-out team-level observations.}
\label{tab:app_nba_calibration}
\begin{tabular}{lrrrr}
\toprule
\textbf{Probability} &
\textbf{Brier} $\downarrow$ &
\textbf{Log Loss} $\downarrow$ &
\textbf{ECE} $\downarrow$ &
\textbf{MCE} $\downarrow$ \\
\midrule
Raw market probability
& \textbf{0.203392}
& \textbf{0.591799}
& \textbf{0.027452}
& \textbf{0.084497} \\
Isotonic calibrated
& 0.204437
& 0.594023
& 0.034241
& 0.500000 \\
\bottomrule
\end{tabular}
\end{table}

These results reinforce the observation in the main text that pre-game
Polymarket prices themselves provide a strong probabilistic benchmark:
a flexible monotonic recalibration does not improve the held-out probability
scores. This application therefore demonstrates that the execution dataset
can recover empirically meaningful market-implied probabilities at scale
and evaluate their calibration against realized domain outcomes.

\subsection{Market-Implied CPI Expectations}
\label{app:cpi}

\paragraph{Data and market reconstruction.}
The CPI application contains 76 Polymarket contracts and
150,981 normalized executions, with each execution uniquely identified by
its transaction hash and log index. The observed execution period extends
from January 16, 2025 to March 23, 2026. Related bucket markets are grouped
by CPI target month, yielding 15 target months in the reconstructed panel,
of which 13 have corresponding realized monthly CPI values and are used in
the quantitative evaluation.

For a trade associated with CPI bucket \(j\), the execution is first
converted to a probability for the positive bucket event,

\[
q_i =
\begin{cases}
p_i, & \text{if the executed outcome represents the bucket event},\\
1-p_i, & \text{if it represents the complementary outcome}.
\end{cases}
\]

Within each hourly interval \(t\), we aggregate executions for bucket \(j\)
using traded share quantity,

\[
\bar q_{j,t}
=
\frac{
\sum_{i\in(j,t)} s_i q_i
}{
\sum_{i\in(j,t)} s_i
}.
\]

The latest available value is carried forward across hours without a new
execution. Probabilities across the contemporaneously available CPI buckets
are then normalized,

\[
\widetilde q_{j,t}
=
\frac{\bar q_{j,t}}
{\sum_k \bar q_{k,t}},
\]

and converted into a continuous market-implied CPI estimate,

\[
\widehat{\mathrm{CPI}}_t
=
\sum_j \widetilde q_{j,t} x_j,
\]

where \(x_j\) denotes the representative CPI value associated with bucket
\(j\). This is the reconstruction used for the size-weighted trajectories
shown in Figure~\ref{fig:cpi_size_weighted_three_months}.

\paragraph{Quantitative comparison.}
We compare the reconstructed market expectation with the Cleveland Fed
inflation nowcast and realized monthly CPI. Across the 13 evaluated target
months, the size-weighted market reconstruction provides 11,481 valid
error observations and achieves an MAE of \(0.0690\) CPI percentage points
and an RMSE of \(0.0850\) percentage points. The corresponding Cleveland
nowcast has an MAE of \(0.0556\) and an RMSE of \(0.0726\) on its available
observations.

\begin{table}[t]
\centering
\small
\caption{Aggregate CPI reconstruction results across 13 target months.
Errors are reported in CPI percentage points.}
\label{tab:app_cpi_results}
\begin{tabular}{lrrr}
\toprule
\textbf{Series} &
\textbf{Observations} &
\textbf{MAE} $\downarrow$ &
\textbf{RMSE} $\downarrow$ \\
\midrule
Polymarket size-weighted
& 11,481 & 0.0690 & 0.0850 \\
Cleveland Fed nowcast
& 9,106 & 0.0556 & 0.0726 \\
\bottomrule
\end{tabular}
\end{table}

Because the two series are not available at exactly the same timestamps,
we additionally compare them on jointly available observations. The
representative months shown in the main text illustrate heterogeneous
relative performance. In December 2025, Polymarket achieves an MAE of
\(0.03795\) percentage points compared with \(0.05834\) for the Cleveland
nowcast. In February 2026, the corresponding values are \(0.01506\) and
\(0.02484\). In January 2026, Cleveland performs better, with an MAE of
\(0.04267\) compared with \(0.07244\) for Polymarket.

\begin{table}[t]
\centering
\small
\caption{Comparison for the three target months visualized in the main text.
MAE is measured on timestamps at which both series are available.}
\label{tab:app_cpi_months}
\begin{tabular}{lrrr}
\toprule
\textbf{Target month} &
\textbf{Common points} &
\textbf{Polymarket MAE} &
\textbf{Cleveland MAE} \\
\midrule
2025-12 & 816 & \textbf{0.03795} & 0.05834 \\
2026-01 & 945 & 0.07244 & \textbf{0.04267} \\
2026-02 & 536 & \textbf{0.01506} & 0.02484 \\
\bottomrule
\end{tabular}
\end{table}

The purpose of this comparison is not to establish uniform dominance over
an established macroeconomic nowcast. Rather, it demonstrates that a set of
discrete prediction-market contracts can be reconstructed into a
high-frequency quantitative signal expressed in the same units as the
eventual economic release, providing a complementary view of evolving
inflation expectations.

\subsection{Resolution Prediction from Partial Market Histories}
\label{app:resolution_prediction}

\paragraph{Market selection and lifecycle samples.}
We construct a cross-domain benchmark from 120 resolved binary markets.
The selected pool spans 12 domains---crypto, esports, finance, games,
geopolitics, movies, politics, pop culture, sports, technology, tennis,
and weather---with 10 markets from each domain. The collection is balanced
at the market level with 60 YES and 60 NO final resolutions and contains
2,427,878 raw execution rows. After validity and lifecycle filtering,
2,427,059 executions remain; the 819 removed observations occur after
final settlement.

For each market with creation time \(t_c\) and settlement time \(t_s\),
we construct three retrospective lifecycle snapshots,

\[
\tau_{\mathrm{early}}
=
t_c + 0.2(t_s-t_c),
\]

\[
\tau_{\mathrm{middle}}
=
t_c + 0.5(t_s-t_c),
\]

and

\[
\tau_{\mathrm{pre}}
=
\max(t_c,t_s-24\mathrm{h}).
\]

Thus, the 120 markets generate 360 market-stage examples. Importantly,
the \(20\%\) and \(50\%\) stages refer to fractions of elapsed time from
market creation to settlement rather than fractions of the observed trade
count.

Markets are ordered by settlement time and split at the market level so
that all three lifecycle observations for the same market remain in the
same partition. Table~\ref{tab:app_supervised_split} summarizes the split.

\begin{table}[t]
\centering
\small
\caption{Market-level chronological split for the resolution-prediction
application.}
\label{tab:app_supervised_split}
\begin{tabular}{lrrr}
\toprule
\textbf{Split} &
\textbf{Markets} &
\textbf{Examples} &
\textbf{YES / NO} \\
\midrule
Train      & 84 & 252 & 44 / 40 \\
Validation & 18 & 54  & 10 / 8 \\
Test       & 18 & 54  & 6 / 12 \\
\midrule
Total      & 120 & 360 & 60 / 60 \\
\bottomrule
\end{tabular}
\end{table}

\paragraph{Input representations.}
The benchmark exposes several representations of the same underlying
execution histories. A tabular representation summarizes each example
using 30 numerical features, including latest YES price, short-horizon
average prices, VWAP, trading volume, trade count, order-flow imbalance,
outcome-volume imbalance, volatility, large-trade fractions, and
missingness measures. A lifecycle-grid representation maps the observed
history into 64 normalized intervals, while an hourly representation
contains the latest 168 hourly observations. Finally, an event-level
representation preserves the most recent 256 executions and directly
encodes price, trade size, outcome, side, time gaps, and missingness.

\begin{table}[t]
\centering
\small
\caption{Input representations constructed from the same market lifecycle
records.}
\label{tab:app_supervised_repr}
\begin{tabular}{lll}
\toprule
\textbf{Representation} &
\textbf{Length} &
\textbf{Coverage} \\
\midrule
Tabular & 1 row & Aggregated market features \\
Hourly grid & 168 & Previous 168 hours \\
Lifecycle grid & 64 & Creation to cutoff \\
Event sequence & 256 & Most recent executions \\
\bottomrule
\end{tabular}
\end{table}

This representation suite allows models with substantially different
inductive biases to be evaluated on the same underlying prediction-market
events: conventional tabular learners operate on aggregated statistics,
regular time-series models operate on discretized trajectories, and
event-sequence models retain the irregular arrival structure of individual
fills.

\paragraph{Models and results.}
We evaluate Last Price as the native prediction-market baseline together
with Random Forest, MLP, TabICLv2, Event GRU, PatchTST, MantisV2,
MiniROCKET, TimesNet, and Neural CDE. Table~\ref{tab:app_supervised_full}
reports the complete point estimates on the 54 test-stage observations
from 18 held-out markets.

\begin{table*}[t]
\centering
\small
\setlength{\tabcolsep}{4.5pt}
\caption{Resolution-prediction performance on the held-out test markets,
pooled across the three lifecycle stages.}
\label{tab:app_supervised_full}
\begin{tabular}{lrrrrrr}
\toprule
\textbf{Model} &
\textbf{Accuracy} $\uparrow$ &
\textbf{AUROC} $\uparrow$ &
\textbf{AUPRC} $\uparrow$ &
\textbf{Log Loss} $\downarrow$ &
\textbf{Brier} $\downarrow$ &
\textbf{ECE} $\downarrow$ \\
\midrule
Last Price
& 0.704 & 0.804 & 0.681
& \textbf{0.474} & \textbf{0.162} & \textbf{0.078} \\
Random Forest
& 0.741 & 0.764 & 0.642
& 0.522 & 0.176 & 0.097 \\
MLP
& 0.759 & \textbf{0.827} & 0.738
& 0.523 & 0.173 & 0.142 \\
TabICLv2
& 0.593 & 0.620 & 0.493
& 0.779 & 0.261 & 0.251 \\
Event GRU
& \textbf{0.796} & 0.814 & \textbf{0.758}
& 0.588 & 0.199 & 0.218 \\
PatchTST
& 0.704 & 0.752 & 0.562
& 0.567 & 0.193 & 0.091 \\
MantisV2
& 0.648 & 0.711 & 0.511
& 0.651 & 0.229 & 0.146 \\
MiniROCKET
& 0.648 & 0.610 & 0.438
& 0.640 & 0.225 & 0.112 \\
TimesNet
& 0.685 & 0.576 & 0.457
& 0.657 & 0.222 & 0.105 \\
Neural CDE
& 0.333 & 0.453 & 0.311
& 0.844 & 0.320 & 0.307 \\
\bottomrule
\end{tabular}
\end{table*}

The results expose a useful distinction between classification and
probabilistic prediction. Event GRU obtains the highest accuracy
(\(0.796\)) and AUPRC (\(0.758\)), while the MLP obtains the highest AUROC
(\(0.827\)). In contrast, the native Last Price baseline achieves the
lowest Log Loss (\(0.474\)), Brier score (\(0.162\)), and ECE (\(0.078\)).
Hence, richer representations can improve particular ranking or
classification objectives, but the contemporaneous market probability
remains a particularly strong baseline for calibrated probabilistic
forecasting.

We further examine performance at different lifecycle stages. The Last
Price Brier score decreases from \(0.2030\) at the early stage to
\(0.1698\) at the middle stage and \(0.1138\) during the final 24 hours,
while its AUROC rises from \(0.639\) to \(0.799\) and \(0.931\),
respectively. This progression provides a direct empirical illustration
of information accumulation: as a market approaches resolution, its
observed trading price becomes increasingly informative about the
eventual outcome.


\section{Polymonitor Panel Inventory and Multi-Agent Analysis}
\label{app:polymonitor_full}

Polymonitor provides 72 registered panels across 11 functional groups,
including three MAS views.  The panels combine prediction-market state
with execution, oracle, news, macroeconomic, weather, sports, and
cross-asset evidence.  Table entries below describe implemented
functions rather than guarantees of freshness or complete coverage.

\subsection{Panel Inventory}
\label{app:polymonitor_panels}

\begin{longtable}{@{}p{0.18\textwidth}p{0.29\textwidth}p{0.47\textwidth}@{}}
\caption{Panels of Polymonitor.}
\label{tab:polymonitor_panels}\\
\toprule
\textbf{Category} & \textbf{Panel} & \textbf{Role} \\
\midrule
\endfirsthead

\toprule
\textbf{Category} & \textbf{Panel} & \textbf{Role} \\
\midrule
\endhead

\bottomrule
\endfoot

\bottomrule
\endlastfoot

\multicolumn{3}{@{}l}{\textbf{Lifecycle}}\\
\midrule
Lifecycle & Active Markets & Market/event navigation and activity. \\
Lifecycle & Orderfilled Flow & Focused-market Polygon executions. \\
Lifecycle & Oracle Feed & Oracle and resolution lifecycle. \\
Lifecycle & Market Summary & Identity, timing, and price summary. \\
Lifecycle & Market Context & Contract rules and resolution context. \\
Lifecycle & Price Surface & Historical probability path. \\
Lifecycle & LOB Depth & Observed YES/NO order-book depth. \\

\midrule
\multicolumn{3}{@{}l}{\textbf{Funding \& Flow}}\\
\midrule
Funding \& Flow & Alpha Signal & Heuristic market-screening signals. \\
Funding \& Flow & PolySignal & Wallet-flow and market-context briefs. \\
Funding \& Flow & Whale Tracker & Large observed transactions. \\
Funding \& Flow & Flow Watch & Unusual or oracle-adjacent flow. \\
Funding \& Flow & New Market Signals & Locally first-seen markets. \\

\midrule
\multicolumn{3}{@{}l}{\textbf{MAS Views}}\\
\midrule
MAS Views & AI Market Brief & MAS overview snapshot. \\
MAS Views & AI Special Markets & MAS special snapshot. \\
MAS Views & AI Trend Radar & MAS trend snapshot. \\

\midrule
\multicolumn{3}{@{}l}{\textbf{News \& Multimedia}}\\
\midrule
News \& Multimedia & Related News & Market-linked news and reports. \\
News \& Multimedia & Breaking Event Radar & Event and attention signals. \\
News \& Multimedia & Market TV Wire & Relevant television/video sources. \\
News \& Multimedia & YouTube Market TV & Curated channels and video feeds. \\
News \& Multimedia & Jin10 Flash & Timestamped macro flash news. \\
News \& Multimedia & BWE News & BWENews RSS feed. \\

\midrule
\multicolumn{3}{@{}l}{\textbf{Crypto \& DeFi}}\\
\midrule
Crypto \& DeFi & Crypto Watch & Spot prices and short-term moves. \\
Crypto \& DeFi & Crypto Funding Watch & Cross-venue perpetual funding. \\
Crypto \& DeFi & DeFi Token Watch & Selected DeFi token prices. \\
Crypto \& DeFi & DeFi Yields & Protocol APY and TVL. \\
Crypto \& DeFi & DeFi Security & Exploit and risk headlines. \\
Crypto \& DeFi & Crypto Perps & Compact perpetual-funding ranking. \\
Crypto \& DeFi & Fear \& Greed & Sentiment and risk gauges. \\
Crypto \& DeFi & Crypto ETF & BTC/ETH ETF flow proxies. \\
Crypto \& DeFi & Stablecoins & Peg and supply observations. \\
Crypto \& DeFi & Chain Policy & Crypto regulation and policy news. \\

\midrule
\multicolumn{3}{@{}l}{\textbf{Finance \& Commodities}}\\
\midrule
Finance \& Commodities & Commodities Watch & Commodity prices and trends. \\
Finance \& Commodities & Commodity Equity Transmission & Commodity--equity exposure map. \\
Finance \& Commodities & TradFi Perps & Tokenized TradFi derivatives. \\
Finance \& Commodities & IPO News & IPO/listing catalysts. \\
Finance \& Commodities & Broker Research & Available analyst research. \\
Finance \& Commodities & Global Indices & Global equity-index context. \\

\midrule
\multicolumn{3}{@{}l}{\textbf{Technology}}\\
\midrule
Technology & AI Model Race & Model releases, benchmarks, and usage. \\
Technology & Big Tech Cap & Technology-company market caps. \\
Technology & App Pulse & App rankings and product news. \\

\midrule
\multicolumn{3}{@{}l}{\textbf{Geopolitics \& Transport}}\\
\midrule
Geopolitics \& Transport & Trade Policy & Curated trade-policy watchlist. \\
Geopolitics \& Transport & Geo/Sanctions Shock & Conflict and sanctions context. \\
Geopolitics \& Transport & Global Transport / Shipping & Aviation/shipping observations. \\

\midrule
\multicolumn{3}{@{}l}{\textbf{Macroeconomics \& CPI}}\\
\midrule
Macroeconomics \& CPI & CPI Release Command Center & Releases, history, and nowcasts. \\
Macroeconomics \& CPI & CPI Components Registry & CPI component-pressure registry. \\
Macroeconomics \& CPI & Goods / Tariff / Supply Watch & Goods and supply-chain pressure. \\
Macroeconomics \& CPI & Labor / Services Monitor & Labor and services inflation. \\
Macroeconomics \& CPI & Fed / Growth Risk Board & Rates and growth indicators. \\
Macroeconomics \& CPI & Polymarket Macro Map & Macro-related contract clusters. \\
Macroeconomics \& CPI & CPI Calendar \& Baseline & Release timing and PM baseline. \\
Macroeconomics \& CPI & Energy / Gasoline Shock & Energy inflation inputs. \\
Macroeconomics \& CPI & Food / Retail Basket & Food CPI pressure. \\
Macroeconomics \& CPI & Supply Tariff Import & Import and tariff pressure. \\
Macroeconomics \& CPI & Shelter Rent OER & Shelter and OER series. \\
Macroeconomics \& CPI & Labor Wage Services & Wage/services pressure. \\
Macroeconomics \& CPI & Growth Demand Recession & Growth and demand indicators. \\
Macroeconomics \& CPI & Inflation Nowcast & Cleveland Fed CPI/PCE nowcast. \\
Macroeconomics \& CPI & Fed Rates PM Gap & Rates versus PM contracts. \\

\midrule
\multicolumn{3}{@{}l}{\textbf{Weather}}\\
\midrule
Weather & Global Temp Monitor & City weather and market coverage. \\
Weather & Weather Market Browser & Weather-contract navigation. \\
Weather & Weather City Snapshot & Selected-city summary. \\
Weather & Weather Quote Curve & Temperature-bin CLOB curve. \\
Weather & Weather Quote Table & Temperature-bin quote table. \\
Weather & WU 1 Day & One-day temperature path. \\
Weather & WU 7 Day & Seven-day forecast. \\
Weather & Weather News & Weather-related news. \\
Weather & World Clock & Local-time utility. \\

\midrule
\multicolumn{3}{@{}l}{\textbf{Sports}}\\
\midrule
Sports & NBA Scoreboard & NBA schedules and scores. \\
Sports & NBA Intel & NBA news and lineup context. \\
Sports & ESPN Matchup Predictor & Available ESPN/BPI forecasts. \\
Sports & Esports Intel & GRID series and PM matching. \\
Sports & Sports Odds & Bookmaker consensus and dispersion. \\

\end{longtable}

\subsection{Multi-Agent System}
\label{app:mas_architecture}

The MAS is a sequential evidence-integration graph with five
deterministic stages and five LLM calls.  Each run produces one
lens-specific snapshot for the three MAS views.

\begin{center}
\small
\texttt{Evidence}
$\rightarrow$
\texttt{Related}
$\rightarrow$
\texttt{Quant}
$\rightarrow$
\texttt{Memory}
$\rightarrow$
\texttt{Microstructure}
$\rightarrow$
\texttt{Catalyst}
$\rightarrow$
\texttt{Resolution}
$\rightarrow$
\texttt{Calibration}
$\rightarrow$
\texttt{Skeptic}
$\rightarrow$
\texttt{Writer}
$\rightarrow$
\texttt{Snapshot}
\end{center}

\begin{table*}[!htbp]
\centering

\MiniTable{Execution Graph}{Stage}{Implementation}{
Evidence / Related / Quant / Memory
& Deterministic Python stages. \\
Microstructure / Catalyst / Resolution
& Specialist LLM calls. \\
Calibration
& Deterministic confidence rules. \\
Skeptic
& LLM evidence challenge. \\
Writer
& LLM final payload synthesis. \\
}
\hfill
\MiniTable{Qwen Run}{Field}{Observed}{
Model & Qwen3.8-27B. \\
Lens & \texttt{special}. \\
Temperature & 0.2. \\
Graph nodes & 10/10 completed. \\
LLM calls & 5/5 completed. \\
Inference & 245.062 s. \\
Tokens & 34,193 in / 4,796 out. \\
Memory & 0 prior episodes. \\
External search & Not executed. \\
Production write & No. \\
}
\hfill
\MiniTable{Context Compression}{Input}{Seed $\rightarrow$ Context}{
Markets & 44 $\rightarrow$ 4. \\
Market groups & 60 $\rightarrow$ 3. \\
Fill tapes & 8 $\rightarrow$ 3. \\
Trades & 24 $\rightarrow$ 0. \\
Oracle events & 24 $\rightarrow$ 0. \\
Content & 12 $\rightarrow$ 4. \\
Alpha signals & 8 $\rightarrow$ 0. \\
Whale signals & 10 $\rightarrow$ 0. \\
Suspicious signals & 10 $\rightarrow$ 0. \\
Derived candidates & -- $\rightarrow$ 8. \\
}

\caption{Compact MAS architecture, audited execution, and model-visible context.}
\label{tab:mas_compact}
\end{table*}

The deterministic related-market stage produces heuristic attention
scores for unusual probability structures; these scores identify cases
for inspection rather than verified arbitrage.  Run IDs, hashes,
node events, model usage, latency, and structured outputs provide
run-level auditability but do not independently validate generated
claims.

\subsection{Actual MAS Prompt Construction}
\label{app:mas_prompts}

The deployed prompts share a common prediction-market policy.
For the four analytical LLM nodes, the actual system message is the
role-specific prefix shown below followed by the shared policy block.
The writer uses its own prefix followed by the same policy.

\begin{promptbox}{Shared Prediction-Market Usefulness Policy}

\textbf{Market-level specificity.}
Do not write generic dashboard observations such as ``activity is
broad'', ``sports supplies count'', or ``liquidity is uneven'' unless
immediately tied to a named market and a price, volume, or trade-count
implication.

\textbf{Preferred thesis form.}
Prefer market-level theses containing a named market, current implied
probability, volume/trade-count, deadline or outcome bucket, catalyst,
and resolution caveat.

\textbf{Probability structure.}
Surface near-50c repricing zones, deadline ladders, term spreads,
mutually related markets, one-sided liquidity, stale prices, and
group-vs-single-market mismatches.

\textbf{Future evidence.}
Explain what observation would move the probability next: an official
source, match result, court or government release, shipping/oil data,
oracle update, or new large fills.

\textbf{Missing evidence.}
Say ``no directional edge'' when price-change data are missing and
identify the missing data rather than filling the gap with category
commentary.

\textbf{Scope.}
Keep claims informational.  Do not recommend trades, position sizing,
or financial advice.

\end{promptbox}

\begin{promptbox}{Microstructure Agent System Prompt}

\textbf{Role.}
You are the market microstructure agent for a Polymarket intelligence
graph. Focus on named markets, implied probability, volume, trade-count,
liquidity concentration, close probabilities, deadline ladders, and
group-vs-single-market mismatches.

\textbf{Composition.}
Append the shared prediction-market usefulness policy above.

\textbf{Return format.}
Return compact JSON only.

\end{promptbox}

\begin{promptbox}{Catalyst Agent System Prompt}

\textbf{Role.}
You are the catalyst research agent for a Polymarket intelligence graph.
Focus on named markets, current prices, concrete external triggers,
related-market catalysts, event timing, and what news or data would move
probability.

\textbf{Composition.}
Append the shared prediction-market usefulness policy above.

\textbf{Return format.}
Return compact JSON only.

\end{promptbox}

\begin{promptbox}{Resolution-Risk Agent System Prompt}

\textbf{Role.}
You are the resolution-risk agent for a Polymarket intelligence graph.
Focus on market wording, deadline buckets, official source hierarchy,
oracle/resolution events, ambiguity, and settlement risk that changes
how prices should be interpreted.

\textbf{Composition.}
Append the shared prediction-market usefulness policy above.

\textbf{Return format.}
Return compact JSON only.

\end{promptbox}

\begin{promptbox}{Skeptic Agent System Prompt}

\textbf{Role.}
You are the skeptic and calibration agent for a Polymarket intelligence
graph. Challenge weak evidence, missing price-change data, stale signals,
narrative overreach, and probability miscalibration.

\textbf{Composition.}
Append the shared prediction-market usefulness policy above.

\textbf{Return format.}
Return compact JSON only.

\end{promptbox}

\begin{promptbox}{Panel Writer System Prompt}

\textbf{Role and objective.}
You are the panel writer for polyData's Forecast Intelligence Graph.
Return compact JSON only. Use the specialist agent outputs as evidence,
but write one coherent dashboard payload.

\textbf{Analytical target.}
The user does not need a category summary. The user needs
prediction-market intelligence: named markets, implied probabilities,
why the market is priced that way, what could move it, and what
resolution wording can break the read.

\textbf{Writing style.}
Write like a prediction-market analyst, not a dashboard narrator.

\textbf{Composition.}
Append the shared prediction-market usefulness policy above.

\end{promptbox}

Specialists return structured findings, risks, watch items, confidence,
and probability adjustments; the writer converts these into the final
brief, focal markets, themes, watchlist, and evidence fields.

\subsection{Audited MAS Output Example}
\label{app:mas_validated_output}

The end-to-end run completed successfully, but the raw writer response
contained arithmetic, freshness, and event-structure errors.  We
therefore report the same case after source-level validation.

\begin{promptbox}{Validated MAS Output from the Audited Run}

\textbf{Observed repricing.}
``US announces end of Iranian blockade by September 30, 2026?''
was recorded at 8.0\%, versus 37.5\% in the reported 24-hour comparison
($-29.5$ points), with approximately \$50.4K volume and 374 trades.

\medskip
\textbf{Related deadlines.}
A separate related-market group contained September 30 at 6.5\% and
December 31 at 65.0\%, a 58.5-point spread.  The 6.5\% group record and
8.0\% single-market record are kept separate.

\medskip
\textbf{Price-source conflict.}
For the Clarity Act market, snapshot, detail, and loaded fill values were
7.2\%, 15.5\%, and approximately 6.2\%, respectively.  The loaded fill
was dated September 19 and is therefore not treated as a September 23
live price.

\medskip
\textbf{Invalid group state.}
The Tampa Bay Rays--Yankees group produced a 299.2\% aggregate with
multiple near-certain child outcomes and was excluded from ordinary
probability-sum interpretation.

\medskip
\textbf{Confidence: Low.}
Eight candidate rows lacked bid/ask observations, detailed trade/oracle
rows were removed during context compression, and no prior-memory or
external-search evidence was available.

\end{promptbox}

\section{ETHICAL STATEMENT}
This research does not involve any human participants. We utilized only open-source models in accordance with their respective licenses, strictly for academic research. Our methodology presents no substantial risks to privacy or security.

\section{Use of Large Language Models}
After the initial draft is completed, we will use specific words to have the LLM refine the text.

\end{document}

%% file: math_commands.tex
\usepackage{amsmath,amsfonts,bm}

\def\eqref#1{equation~\ref{#1}}

\def\1{\bm{1}}

\DeclareMathAlphabet{\mathsfit}{\encodingdefault}{\sfdefault}{m}{sl}
\SetMathAlphabet{\mathsfit}{bold}{\encodingdefault}{\sfdefault}{bx}{n}



%% file: sample-base.bib
@String{Computer = "{IEEE} Computer" }

@String{Springer = "Springer-Verlag" }

@article{saguillo2025unravelling,
  title={Unravelling the Probabilistic Forest: Arbitrage in Prediction Markets},
  author={Saguillo, Oriol and Ghafouri, Vahid and Kiffer, Lucianna and Suarez-Tangil, Guillermo},
  journal={arXiv preprint arXiv:2508.03474},
  year={2025}
}

@article{reichenbach2025exploring,
  title={Exploring Decentralized Prediction Markets: Accuracy, Skill, and Bias on Polymarket},
  author={Reichenbach, Felix and Walther, Martin},
  journal={Skill, and Bias on Polymarket (December 12, 2025)},
  year={2025}
}

@article{cutting2025betting,
  title={Are Betting Markets Better than Polling in Predicting Political Elections?},
  author={Cutting, Laurie E and Hughes-Berheim, Sarah S and Johnson, Paul M and Baroud, Hiba and Goldstein, Brett},
  journal={arXiv preprint arXiv:2507.08921},
  year={2025}
}

@article{capponi2025semantic,
  title={Semantic Trading: Agentic AI for Clustering and Relationship Discovery in Prediction Markets},
  author={Capponi, Agostino and Gliozzo, Alfio and Zhu, Brian},
  journal={arXiv preprint arXiv:2512.02436},
  year={2025}
}

@article{becker2026political,
  title={Political Uncertainty and Credit Risk: The Role of Event Markets in Forecasting Ukraine's Sovereign Spreads},
  author={Becker, Mary and McGurk, Zachary},
  journal={Scottish Journal of Political Economy},
  pages={e70062},
  year={2026},
  publisher={Wiley Online Library}
}

@article{ng2026price,
  title={Price Discovery and Trading in Modern Prediction Markets},
  author={Ng, Hunter and Peng, Lin and Tao, Yubo and Zhou, Dexin},
  journal={Available at SSRN 5331995},
  year={2026}
}

@techreport{diercks2026kalshi,
  title={Kalshi and the Rise of Macro Markets},
  author={Diercks, Anthony M and Katz, Jared Dean and Wright, Jonathan H},
  year={2026},
  institution={National Bureau of Economic Research}
}

@article{eichengreen2025under,
  title={Under pressure? Central bank independence meets blockchain prediction markets},
  author={Eichengreen, Barry and Viswanath-Natraj, Ganesh and Wang, Junxuan and Wang, Zijie},
  journal={Central Bank Independence Meets Blockchain Prediction Markets (July 26, 2025)},
  year={2025}
}

@article{rahman2025sok,
  title={SoK: Market Microstructure for Decentralized Prediction Markets (DePMs)},
  author={Rahman, Nahid and Al-Chami, Joseph and Clark, Jeremy},
  journal={arXiv preprint arXiv:2510.15612},
  year={2025}
}

@article{luo2024multi,
  title={Multi-chain graphs of graphs: A new approach to analyzing blockchain datasets},
  author={Luo, Bingqiao and Zhang, Zhen and Wang, Qian and He, Bingsheng},
  journal={Advances in Neural Information Processing Systems},
  volume={37},
  pages={28490--28514},
  year={2024}
}

@inproceedings{zhou2025graph,
  title={Graph neural networks for bridge swap link prediction in Uniswap v3},
  author={Zhou, Qingran and Liu, Eric and Brini, Alessio},
  booktitle={Proceedings of the 6th ACM International Conference on AI in Finance},
  pages={247--255},
  year={2025}
}

@inproceedings{ni2024money,
  title={Money never sleeps: Maximizing liquidity mining yields in decentralized finance},
  author={Ni, Wangze and Yiwei, Zhao and Sun, Weijie and Chen, Lei and Cheng, Peng and Zhang, Chen Jason and Lin, Xuemin},
  booktitle={Proceedings of the 30th ACM SIGKDD Conference on Knowledge Discovery and Data Mining},
  pages={2248--2259},
  year={2024}
}

@inproceedings{guan2024characterizing,
  title={Characterizing Ethereum address poisoning attack},
  author={Guan, Shixuan and Li, Kai},
  booktitle={Proceedings of the 2024 on ACM SIGSAC Conference on Computer and Communications Security},
  pages={986--1000},
  year={2024}
}

@inproceedings{cernera2023token,
  title={Token spammers, rug pulls, and sniper bots: An analysis of the ecosystem of tokens in ethereum and in the binance smart chain ($\{$$\{$$\{$$\{$$\{$BNB$\}$$\}$$\}$$\}$$\}$)},
  author={Cernera, Federico and La Morgia, Massimo and Mei, Alessandro and Sassi, Francesco},
  booktitle={32nd USENIX security symposium (USENIX security 23)},
  pages={3349--3366},
  year={2023}
}

@inproceedings{wang2024nft1000,
  title={Nft1000: A cross-modal dataset for non-fungible token retrieval},
  author={Wang, Shuxun and Lei, Yunfei and Zhang, Ziqi and Liu, Wei and Liu, Haowei and Yang, Li and Li, Bing and Li, Wenjuan and Gao, Jin and Hu, Weiming},
  booktitle={Proceedings of the 32nd ACM International Conference on Multimedia},
  pages={2214--2222},
  year={2024}
}

@inproceedings{zhou2024stop,
  title={Stop pulling my rug: Exposing rug pull risks in crypto token to investors},
  author={Zhou, Yuanhang and Sun, Jingxuan and Ma, Fuchen and Chen, Yuanliang and Yan, Zhen and Jiang, Yu},
  booktitle={Proceedings of the 46th International Conference on Software Engineering: Software Engineering in Practice},
  pages={228--239},
  year={2024}
}

@misc{clevelandfed_inflation_nowcast,
  author       = {{Federal Reserve Bank of Cleveland}},
  title        = {Inflation Nowcasting},
  howpublished = {\url{https://www.clevelandfed.org/indicators-and-data/inflation-nowcasting}},
  year         = {2026},
  doi          = {10.26509/frbc-inflationnowcast},
  note         = {Center for Inflation Research; updated each business day. Accessed: 2026-04-07}
}

@article{cong2025primer,
  title={A primer on oracle economics},
  author={Cong, Lin William and Fox, Liam and Li, Siguang and Zhou, Luofeng},
  journal={Journal of Corporate Finance},
  volume={94},
  pages={102800},
  year={2025},
  publisher={Elsevier}
}

@article{cong2023onchain,
  title={Onchain Reputation: Introducing Dynamic Incentives into Distributed Networks},
  author={Cong, Lin William and Zhou, Luofeng},
  journal={Available at SSRN 4540862},
  year={2023}
}

@techreport{cong2025financial,
  title={Financial and informational integration through oracle networks},
  author={Cong, Lin William and Prasad, Eswar S and Rabetti, Daniel},
  year={2025},
  institution={National Bureau of Economic Research}
}

@incollection{chen2021brief,
  title={A brief introduction to blockchain economics},
  author={Chen, Long and Cong, Lin William and Xiao, Yizhou},
  booktitle={Information for efficient decision making: Big data, blockchain and relevance},
  pages={1--40},
  year={2021},
  publisher={World Scientific}
}

@article{sun2025sok,
  title={Sok: A taxonomic analysis of defi rug pulls: Types, dataset, and tool assessment},
  author={Sun, Dianxiang and Ma, Wei and Nie, Liming and Liu, Yang},
  journal={Proceedings of the ACM on Software Engineering},
  volume={2},
  number={ISSTA},
  pages={550--572},
  year={2025},
  publisher={ACM New York, NY, USA}
}

@article{wang2023ex,
  title={Ex-graph: A pioneering dataset bridging ethereum and x},
  author={Wang, Qian and Zhang, Zhen and Liu, Zemin and Lu, Shengliang and Luo, Bingqiao and He, Bingsheng},
  journal={arXiv preprint arXiv:2310.01015},
  year={2023}
}

@article{mongardini2025midsummer,
  title={A Midsummer Meme's Dream: Investigating Market Manipulations in the Meme Coin Ecosystem},
  author={Mongardini, Alberto Maria and Mei, Alessandro},
  journal={arXiv preprint arXiv:2507.01963},
  year={2025}
}

@inproceedings{guan2025security,
  title={Security perceptions of users in stablecoins: Advantages and risks within the cryptocurrency ecosystem},
  author={Guan, Maggie Yongqi and Yu, Yaman and Sharma, Tanusree and Huang, Molly Zhuangtong and Qin, Kaihua and Wang, Yang and Wang, Kanye Ye},
  booktitle={2025 IEEE Symposium on Security and Privacy (SP)},
  pages={2753--2771},
  year={2025},
  organization={IEEE}
}

@article{cao2026price,
  title={The Price of Interoperability: Exploring Cross-Chain Bridges and Their Economic Consequences},
  author={Cao, Yiyue and Zheng, Mingzhe and Cong, Lin William and Li, Siguang and Wang, Xuechao},
  journal={arXiv preprint arXiv:2604.03083},
  year={2026}
}

@inproceedings{gramoli2023diablo,
  title={Diablo: A benchmark suite for blockchains},
  author={Gramoli, Vincent and Guerraoui, Rachid and Lebedev, Andrei and Natoli, Chris and Voron, Gauthier},
  booktitle={Proceedings of the Eighteenth European Conference on Computer Systems},
  pages={540--556},
  year={2023}
}

@inproceedings{hu2024zipzap,
  title={Zipzap: Efficient training of language models for large-scale fraud detection on blockchain},
  author={Hu, Sihao and Huang, Tiansheng and Chow, Ka-Ho and Wei, Wenqi and Wu, Yanzhao and Liu, Ling},
  booktitle={Proceedings of the ACM Web Conference 2024},
  pages={2807--2816},
  year={2024}
}

@inproceedings{zhou2023sok,
  title={Sok: Decentralized finance (defi) attacks},
  author={Zhou, Liyi and Xiong, Xihan and Ernstberger, Jens and Chaliasos, Stefanos and Wang, Zhipeng and Wang, Ye and Qin, Kaihua and Wattenhofer, Roger and Song, Dawn and Gervais, Arthur},
  booktitle={2023 IEEE Symposium on Security and Privacy (SP)},
  pages={2444--2461},
  year={2023},
  organization={IEEE}
}

@misc{uma_protocol,
  author       = {{Risk Labs Foundation}},
  title        = {{UMA}: A Decentralized Optimistic Oracle},
  year         = {2020},
  howpublished = {\url{https://uma.xyz/}},
  note         = {Accessed: 2026-04-13},
  description  = {Optimistic oracle and dispute arbitration system for bringing arbitrary verifiable data onchain; supports prediction markets, insurance, cross-chain bridges, and RWAs}
}

@article{barlow1972isotonic,
  title={The isotonic regression problem and its dual},
  author={Barlow, Richard E and Brunk, Hugh D},
  journal={Journal of the American Statistical Association},
  volume={67},
  number={337},
  pages={140--147},
  year={1972},
  publisher={Taylor \& Francis}
}

@misc{bls_cpi_release_schedule,
  author       = {{U.S. Bureau of Labor Statistics}},
  title        = {Schedule of Releases for the Consumer Price Index},
  year         = {2026},
  howpublished = {\url{https://www.bls.gov/schedule/news_release/cpi.htm}},
  note         = {Accessed: 2026-04-13}
}

@misc{polymarketGammaAPI2026,
  author       = {{Polymarket}},
  title        = {Discover Markets: Gamma Market Data API},
  year         = {2026},
  howpublished = {Polymarket Documentation},
  url          = {https://docs.polymarket.com/market-data/discover-markets},
  note         = {Accessed September 14, 2026}
}

@misc{polymarketTradingAPI2026,
  author       = {{Polymarket}},
  title        = {Trading API: Place Your First Order},
  year         = {2026},
  howpublished = {Polymarket Documentation},
  url          = {https://docs.polymarket.com/trading/quickstart},
  note         = {Accessed September 14, 2026}
}

@misc{polymarketContracts2026,
  author       = {{Polymarket}},
  title        = {Contracts},
  year         = {2026},
  howpublished = {Polymarket Documentation},
  url          = {https://docs.polymarket.com/resources/contracts},
  note         = {Accessed September 14, 2026}
}

@misc{polymarketGitHub2026,
  author       = {{Polymarket}},
  title        = {Polymarket Official GitHub Organization},
  year         = {2026},
  howpublished = {GitHub},
  url          = {https://github.com/Polymarket},
  note         = {Accessed September 14, 2026}
}

@misc{polymarketUmaCtfAdapter2026,
  author       = {{Polymarket}},
  title        = {Polymarket UMA CTF Adapter},
  year         = {2026},
  howpublished = {GitHub Repository},
  url          = {https://github.com/Polymarket/uma-ctf-adapter},
  note         = {Accessed September 14, 2026}
}

@misc{polygonRPC2026,
  author       = {{Polygon Labs}},
  title        = {Polygon PoS RPC Endpoints},
  year         = {2026},
  howpublished = {Polygon Developer Documentation},
  url          = {https://docs.polygon.technology/pos/reference/rpc-endpoints},
  note         = {Accessed September 14, 2026}
}

@article{breiman2001random,
  title={Random forests},
  author={Breiman, Leo},
  journal={Machine learning},
  volume={45},
  number={1},
  pages={5--32},
  year={2001},
  publisher={Springer}
}

@article{rumelhart1986learning,
  title={Learning representations by back-propagating errors},
  author={Rumelhart, David E and Hinton, Geoffrey E and Williams, Ronald J},
  journal={nature},
  volume={323},
  number={6088},
  pages={533--536},
  year={1986},
  publisher={Nature Publishing Group UK London}
}

@article{qu2026tabiclv2,
  title   = {{TabICLv2}: A Better, Faster, Scalable, and Open Tabular Foundation Model},
  author  = {Qu, Jingang and Holzm{\"u}ller, David and Varoquaux, Ga{\"e}l and Le Morvan, Marine},
  journal = {arXiv preprint arXiv:2602.11139},
  year    = {2026}
}

@inproceedings{cho2014learning,
  title={Learning phrase representations using RNN encoder--decoder for statistical machine translation},
  author={Cho, Kyunghyun and Van Merri{\"e}nboer, Bart and Gul{\c{c}}ehre, {\c{C}}a{\u{g}}lar and Bahdanau, Dzmitry and Bougares, Fethi and Schwenk, Holger and Bengio, Yoshua},
  booktitle={Proceedings of the 2014 conference on empirical methods in natural language processing (EMNLP)},
  pages={1724--1734},
  year={2014}
}

@article{nie2022time,
  title={A time series is worth 64 words: Long-term forecasting with transformers},
  author={Nie, Yuqi and Nguyen, Nam H and Sinthong, Phanwadee and Kalagnanam, Jayant},
  journal={arXiv preprint arXiv:2211.14730},
  year={2022}
}

@article{feofanov2026mantisv2,
  title={Mantisv2: Closing the zero-shot gap in time series classification with synthetic data and test-time strategies},
  author={Feofanov, Vasilii and Wen, Songkang and Zhang, Jianfeng and Pan, Lujia and Redko, Ievgen},
  journal={arXiv preprint arXiv:2602.17868},
  year={2026}
}

@inproceedings{dempster2021minirocket,
  title={Minirocket: A very fast (almost) deterministic transform for time series classification},
  author={Dempster, Angus and Schmidt, Daniel F and Webb, Geoffrey I},
  booktitle={Proceedings of the 27th ACM SIGKDD conference on knowledge discovery \& data mining},
  pages={248--257},
  year={2021}
}

@article{wu2022timesnet,
  title={Timesnet: Temporal 2d-variation modeling for general time series analysis},
  author={Wu, Haixu and Hu, Tengge and Liu, Yong and Zhou, Hang and Wang, Jianmin and Long, Mingsheng},
  journal={arXiv preprint arXiv:2210.02186},
  year={2022}
}

@article{kidger2020neural,
  title={Neural controlled differential equations for irregular time series},
  author={Kidger, Patrick and Morrill, James and Foster, James and Lyons, Terry},
  journal={Advances in neural information processing systems},
  volume={33},
  pages={6696--6707},
  year={2020}
}
